# LEME: Open Large Language Models for Ophthalmology with Advanced Reasoning and Clinical Validation

Hyunjae Kim[1]*, Xuguang Ai[1]*, Sahana Srinivasan[2,3]*, Aidan Gilson[4], Maxwell B. Singer[5], Krithi Pushpanathan[2,3], Qianqian Xie[1], Jungwoo Park[6], Serina Applebaum[5], Gabriel Dawei Yang[7], Minjie Zou[2,3], David Ziyou Chen[2,3,8], Ke Zou[2,3], Soshian Sarrafpour[5], Ji Liu[5], Yu Yin[9], Jimin Huang[1], Quang Ngoc Nguyen[10,11,12,13], Erping Long[14], Peixing Wan[15], Dianbo Liu[2,3], Richard Hintz[16], W. Jim Zheng[17], Sophia Y. Wang[10], Lucila Ohno-Machado[1], Hua Xu[1], Ron A. Adelman[5], Luciano V. Del Priore[5], Yih-Chung Tham[2,3,7,18†], Qingyu Chen[1†]

1. Department of Biomedical Informatics and Data Science, Yale School of Medicine, Yale University, New Haven, CT, USA
2. Centre for Innovation and Precision Eye Health, Department of Ophthalmology, Yong Loo Lin School of Medicine, National University of Singapore, Singapore
3. Department of Ophthalmology, Yong Loo Lin School of Medicine, National University of Singapore, Singapore
4. Department of Ophthalmology, Massachusetts Eye and Ear, Harvard Medical School, Boston, MA, USA
5. Department of Ophthalmology and Visual Science, Yale School of Medicine, Yale University, New Haven, CT, USA
6. Department of Computer Science, Korea University, Seoul, Republic of Korea
7. Singapore Eye Research Institute, Singapore National Eye Centre, Singapore
8. Department of Ophthalmology, National University Hospital, Singapore
9. Department of Computer Science, University of Liverpool, Liverpool, United Kingdom
10. Byers Eye Institute, Stanford University, CA, USA
11. Institute of Ophthalmology, University College London, London, United Kingdom
12. Institute of Health Informatics, University College London, London, United Kingdom
13. Moorfields Eye Hospital, London, United Kingdom
14. Division of Cancer Epidemiology and Genetics, National Cancer Institute, National Institutes of Health, Bethesda, MD, USA

15. Center for Cancer Research, National Cancer Institute, National Institutes of Health, Bethesda, MD, USA

16. Research Informatics Office, Yale School of Medicine, Yale University, New Haven, CT, USA

17. School of Biomedical Informatics, University of Texas Health Science Center, Houston, TX, USA

18. Eye Academic Clinical Program (Eye ACP), Duke-NUS Medical School, Singapore

*Contributed equally as first authors

†Contributed equally as last authors

**Corresponding authors:**

Dr. Yih-Chung Tham

Yong Loo Lin School of Medicine, National University of Singapore.

Level 13, MD1 Tahir Foundation Building, 12 Science Drive 2, Singapore 117549

Tel: +65 65767298, Fax: +65 6225 2568; Email: thamyc@nus.edu.sg

Dr. Qingyu Chen

Yale School of Medicine, Yale University

101 College Street, Fl 10, New Haven, Connecticut, 06510, US

Email: qingyu.chen@yale.edu

## Abstract

The rising prevalence of eye diseases poses a growing public health burden. Large language models (LLMs) offer a promising path to reduce documentation workload and support clinical decision-making. However, few have been tailored for ophthalmology, and most evaluations focus mainly on knowledge-based question answering (QA) without clinically relevant benchmarks or real-world validation. Here, we present LEME, a suite of open-weight LLMs developed through a two-stage process: (1) instruction tuning on 200,000 samples from clinical guidelines, textbooks, and case reports to enhance reasoning and task-following, and (2) reinforcement learning with close to 30,000 preference labels to enhance accuracy and informativeness. LEME was evaluated on five curated zero-shot benchmarks spanning tasks such as patient QA, consultation, and treatment planning. It outperformed all seven baselines (all $p < 0.004$), exceeding GPT-4o by 3.32% (absolute ROUGE-L gain). It was further evaluated on three downstream tasks using deidentified patient data, reviewed by clinicians. In patient QA, LEME received the highest ratings from attending clinicians in 3 out of 4 criteria, with scores of 4.67 for factuality, 4.77 for specificity, 4.79 for completeness, and 4.88 for safety (1–5 scale). Its completeness score surpassed that of expert-written answers (4.79 vs. 4.56; $p = 0.015$). In visual acuity extraction, LEME achieved the highest F1, outperforming LLaMA-3 by 14.1% and Eye-LLaMA by 59.0%. In a pilot evaluation on assessment and treatment planning for diabetic retinopathy, age-related macular degeneration, and glaucoma, LEME received scores of 4.36 for factuality, 4.55 for specificity, 4.42 for completeness, and 4.36 for safety, approaching attending-level performance. All models, data, and code will be released to support further development and clinical translation, laying the groundwork for improved efficiency and patient care.

# Introduction

Large language models (LLMs) demonstrate strong zero- and few-shot learning capabilities, enabling a single model to perform a wide range of general tasks with minimal examples[1-6]. They also exhibit reasoning abilities, including synthesizing information across diverse sources and step-by-step problem solving[7-10]. These capabilities have opened new avenues in medical domains, including text summarization, addressing patient queries, and disease diagnosis[11-13]. Recently, pioneering efforts have explored their use in medical domain where clinical documentation and processes remain a significant burden[14-16]. Clinicians must interpret information from imaging and test reports from patient visits to summaries longitudinal patient histories and draft assessment and treatment plans. In ophthalmology, a specialty characterized by structured, protocol-driven workflows and analysis of imaging reports, LLMs has been reported to leverage LLMs to support clinical decision-making, streamline documentation, enhance patient education, and facilitate large-scale research through the analysis of ophthalmic data[16-19].

However, despite the potential, the development and adoption of LLMs in ophthalmology remain limited due to several key barriers. Proprietary LLMs like GPT-4o[20], o1-mini[21], and Gemini 2.5[22], while representing state-of-the-art general-domain performance, are typically accessed via Application Programming Interfaces (APIs) that require patient data to be transmitted to third-party servers (**Fig. 1a**). This raises security and privacy concerns, particularly in healthcare scenarios where external data sharing is often restricted[23-25]. Several recent studies introduced open-weight LLMs for ophthalmology that can be adapted and deployed locally[26,27]; however, these models have substantial limitations (illustrated in **Fig. 1b**). First, in terms of model availability, only one model, Eye-LLaMA[28], is publicly available. Nevertheless, it was trained using copyrighted data restricted for AI development, leaving its licensing status in question. Second, from a methodological perspective, these models rely on small or earlier-generation architectures and have undergone minimal fine-tuning, lacking reasoning-centric training and preference alignment[29,30]. Most critically, evaluations have focused largely on knowledge-based testing such as multiple-choice QA with few or no demonstrations on real-world clinical tasks.[27,31-37]. Overall, current evaluations lack downstream validation on clinically relevant benchmarks and workflows[38,39].

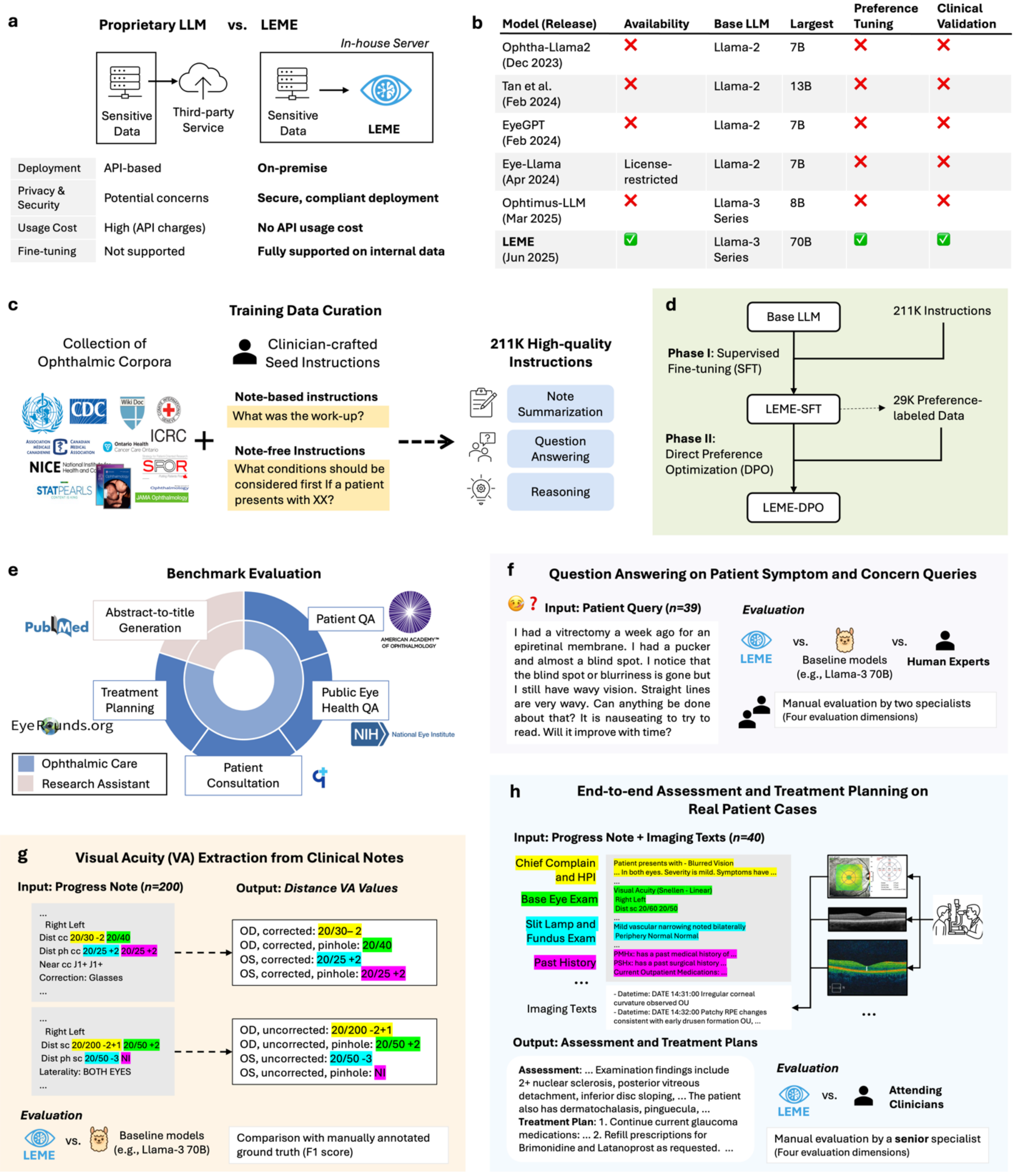

**Fig. 1 | Overview of study design. a,** LEME enables secure, on-premise deployment, addressing privacy and customization limitations of proprietary LLMs. **b,** Compared to prior models, LEME is open-weight, preference-tuned, and clinically validated. **c,** Training data comprises 211K instructions derived from ophthalmology corpora and clinician-authored instructions. **d,** Two-stage training: supervised fine-tuning (SFT) followed by direct preference optimization (DPO). **e,** Evaluation covers five ophthalmology-specific benchmarks, including four clinically relevant tasks and one research assistant task. **f-h,** Downstream validation includes real patient data and manual specialist review.

To address these gaps, we present LEME (**L**anguage-**E**nhanced Foundation **M**odel for **E**ye Care), open-weight LLMs specifically tailored for ophthalmology. LEME is trained based on a two-stage framework, comprising supervised fine-tuning and reinforcement learning (**Fig. 1c,d**). In the first stage, we conducted instruction tuning on 211,149 samples to strengthen clinical reasoning and task adherence. These samples were curated from the most comprehensive ophthalmology corpus to date, which was constructed from clinical guidelines[40], textbooks[41], and case reports[42] freely available for research use, including AI development. In the second stage, we further refined the model using Direct Preference Optimization (DPO)[43] on 29,747 preference-labeled samples. This step guided the model to align outputs with desired qualities of accuracy and informativeness. The data and training pipeline are designed to be sustainable and can be readily adapted to emerging LLM backbones for domain-specific specialization.

We systematically evaluated LEME against seven leading LLMs (GPT-4o, o1-mini, LLaMA-3 70B, LLaMA-2 7B and 70B, Meditron 70B, and Eye-LLaMA 7B), spanning general-purpose (proprietary and open-weight), medical-specific, and ophthalmology-focused models. To go beyond prior work that focused primarily on multiple-choice QA, we curated a benchmark of open-ended, ophthalmology-specific tasks designed to better reflect real-world needs (**Fig. 1e**). These tasks were drawn from publicly available sources, including patient-facing Q&A forums and authoritative resources such as the American Academy of Ophthalmology (AAO) and the National Eye Institute (NEI). Our benchmark emphasizes scenarios central to clinical ophthalmic care and research, including disease-specific question answering, interpretation of examination findings, physician-patient communication, treatment/management planning, and summarizing research abstracts into titles.

Additionally, LEME was evaluated across three clinically grounded downstream tasks (**Fig. 1f-h**). First, to assess its ability to address patient concerns, the model was tested on real-world patient questions related to eye diseases. Responses were independently rated by two ophthalmology specialists for factuality, specificity, completeness, and safety. Second, we examined the model's capacity to extract visual acuity recordings from real-world patient clinical notes, a fundamental yet error-prone task in ophthalmology that requires precise identification and restructuring of information from free-text clinical documentation[44-46]. Third, we conducted a pilot evaluation of automated generation of assessment and treatment plans from progress notes and imaging report texts using randomly selected retrospective patient records with diabetic retinopathy, age-related macular degeneration, or glaucoma. A senior ophthalmologist independently compared the model-generated outputs with the original documentation authored by attending clinicians. This task is labor-intensive in clinical workflows and requires nuanced interpretation and reasoning across diverse information sources, such as chief complaints, medication history, and imaging reports, which is challenging and largely unexplored in the existing evaluations.

Our key contributions are as follows: (1) We present LEME, open-weight LLMs for ophthalmology, trained via reasoning-intensive instruction tuning and preference-based reinforcement learning. (2) We perform clinically relevant evaluations of LLMs in ophthalmology, going beyond multiple-choice QA to include five curated open-ended benchmarks and three real-world downstream tasks using actual patient notes, with expert review by ophthalmology specialists. (3) Leveraging this design, LEME achieves superior performance over strong baseline LLMs (e.g., GPT-4o, LLaMA-3 70B, Eye-LLaMA) and exhibits comparable or superior performance when compared to clinicians in select aspects, such as completeness, highlighting its potential for informative clinical use. (4) All model weights, datasets, and codes will be publicly released to allow secure clinical deployment and support future research in ophthalmic AI.

## Results

### Benchmark evaluation on five open-ended tasks

We evaluated two variants of our LEME model: LEME-SFT, trained with supervised fine-tuning on top of LLaMA-3 70B[47], and LEME-DPO, further refined via Direct Preference Optimization. Both are compared against each other and strong external baselines (GPT-4o[20], o1-mini[21], LLaMA-3 70B, LLaMA-2 7B and 70B[48], Meditron 70B[40], and Eye-LLaMA 7B[28]). Representative examples for each benchmark task are provided in **Extended Data Table 1**.

LEME models deliver superior performance on open-ended tasks (**Fig. 2**). Across every dataset, LEME-SFT attained the highest ROUGE-L and significantly outperform every baseline including GPT-4o and LLaMA-3 70B (Holm-adjusted $p < 0.004$ for each model-dataset comparison).

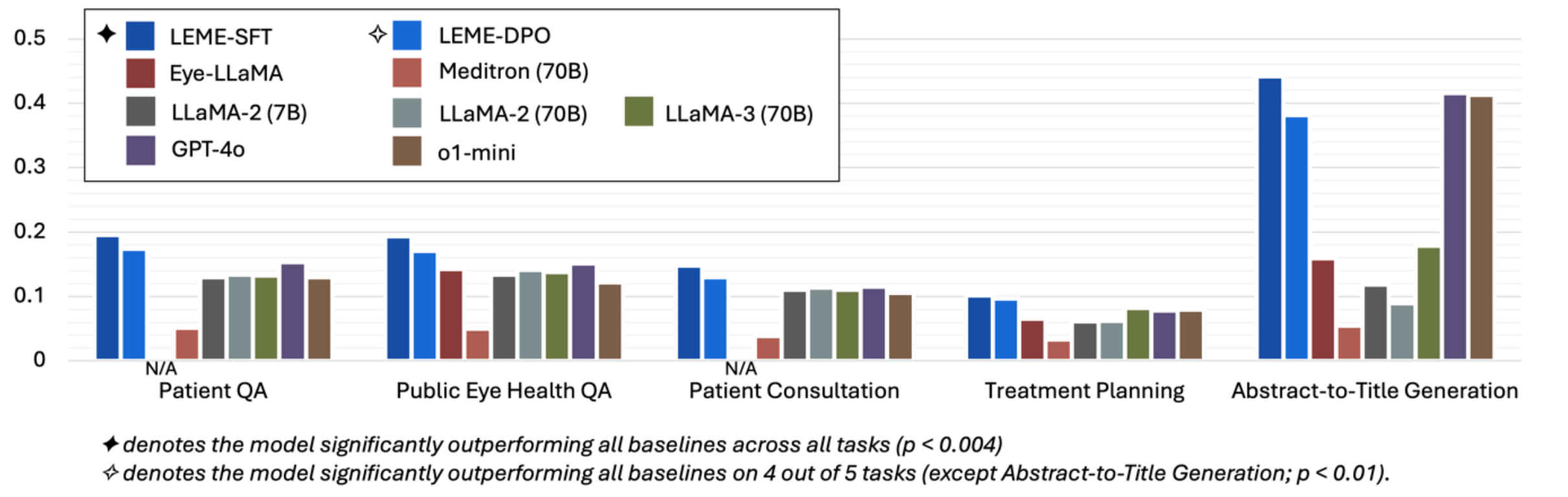

**Fig. 2 | Benchmark evaluation on five open-ended ophthalmology tasks.** Performance is reported using ROUGE-L. LEME-SFT significantly outperformed all baselines across all five tasks (✦) (Benjamini–Hochberg-adjusted $p < 0.004$), while LEME-DPO outperformed all baselines on four of five tasks (✧) ($p < 0.01$). Eye-LLaMA was not evaluated on patient QA and patient consultation due to potential training data overlap.

Against our own LEME-DPO variant, LEME-SFT is significantly better on four datasets ($p < 0.004$) and statistically tied on Treatment Planning ($p = 0.058$). LEME-DPO also leads on four clinically relevant tasks (patient QA, public eye health QA, patient consultation, and treatment planning), outperforming all external baselines in each pairing ($p < 0.01$). On the title generation task, it still surpasses the open-weight baselines but trails GPT-4o and o1-mini. **Extended Data Table 2** presents the full results with 95% confidence intervals. It also includes additional metrics such as BERTScore[49], which support the same conclusion.

## Question answering on patient symptom and concern queries

We evaluated whether LEME can interpret patients' real-world symptoms and concerns, and respond with accurate, complete, and safe guidance. Two attending clinicians independently evaluated responses to 39 patient-centered queries from the American Academy of Ophthalmology. Each query was answered by four response types: LEME-SFT, LEME-DPO, LLaMA-3 70B, and original expert-written answers (Human). To minimize potential bias and prevent identification, all responses were anonymized and randomly shuffled, in accordance with established evaluation practices[50,51]. All responses were rated on a five-point scale across four dimensions: factuality, specificity, completeness, and safety.

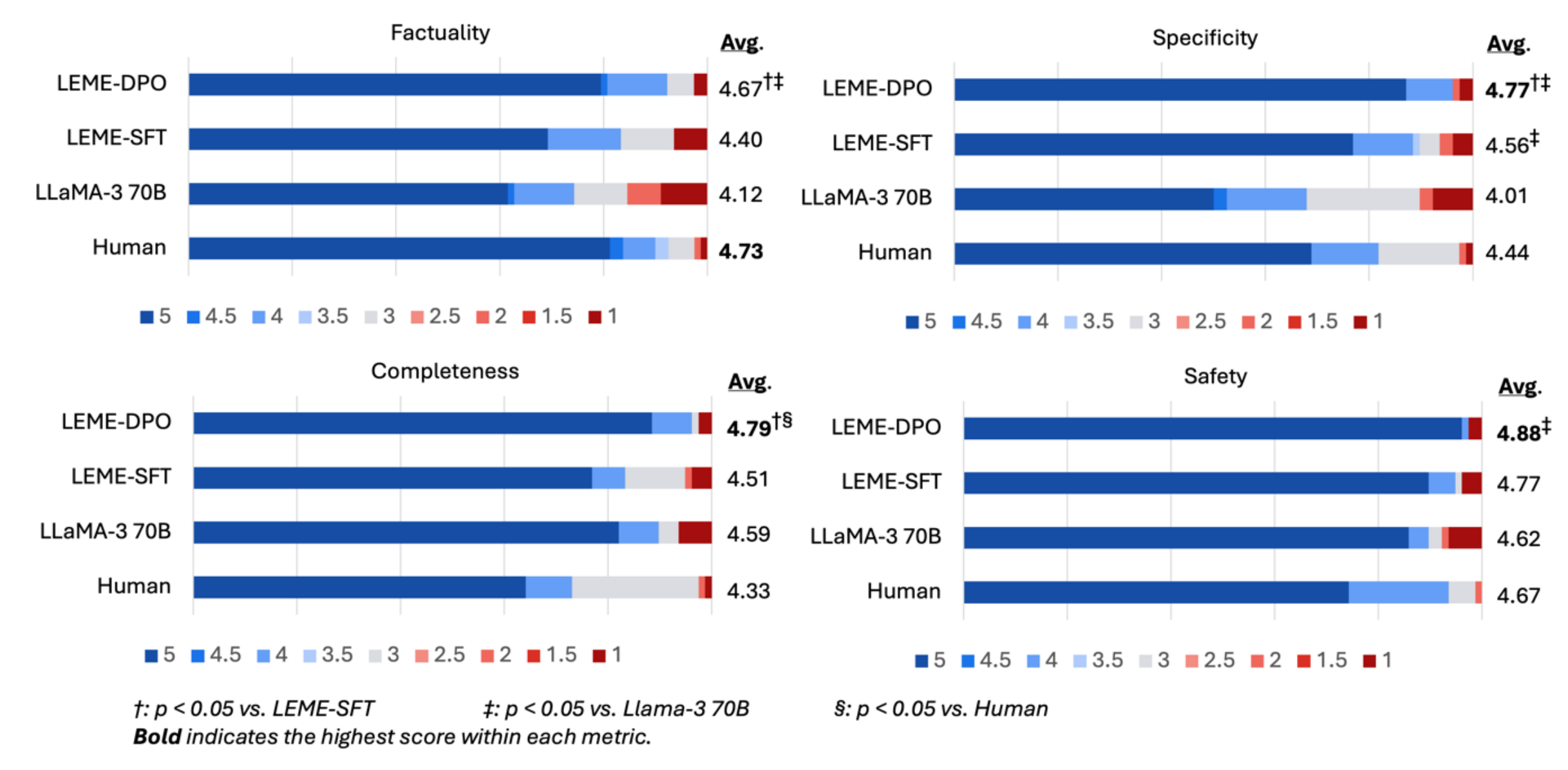

**Fig. 3 | Expert evaluation of responses to 39 patient-authored symptom and concern queries.** Responses were rated by two attending ophthalmologists on a 5-point scale across four dimensions. Human responses scored highest in factuality (4.73), while LEME-DPO achieved the top scores for specificity (4.77, $p < 0.05$ vs. LEME-SFT and LLaMA-3), completeness (4.79, $p < 0.05$ vs. LEME-SFT and Human), and safety (4.88, $p < 0.05$ vs. LLaMA-3).

As shown in **Fig. 3**, LEME-DPO achieved the highest in three out of the four evaluation dimensions and significantly surpassed expert-written answers in completeness. In factuality, LEME-DPO scored 4.67, closely approximating expert responses (4.73) and significantly outperforming LLaMA-3 (4.12) by 0.55. For specificity, LEME-DPO attained the highest score (4.77), exceeding both LEME-SFT (4.56) and LLaMA-3 (4.01). In completeness, LEME-DPO again led with 4.79, showing significant gains over both LEME-SFT (4.51) and expert responses (4.33) by 0.46 points. In safety, it achieved the top score of 4.88, outperforming LLaMA-3 (4.62). All p-values are from pairwise permutation tests, adjusted using the Benjamini–Hochberg procedure (all $p < 0.05$).

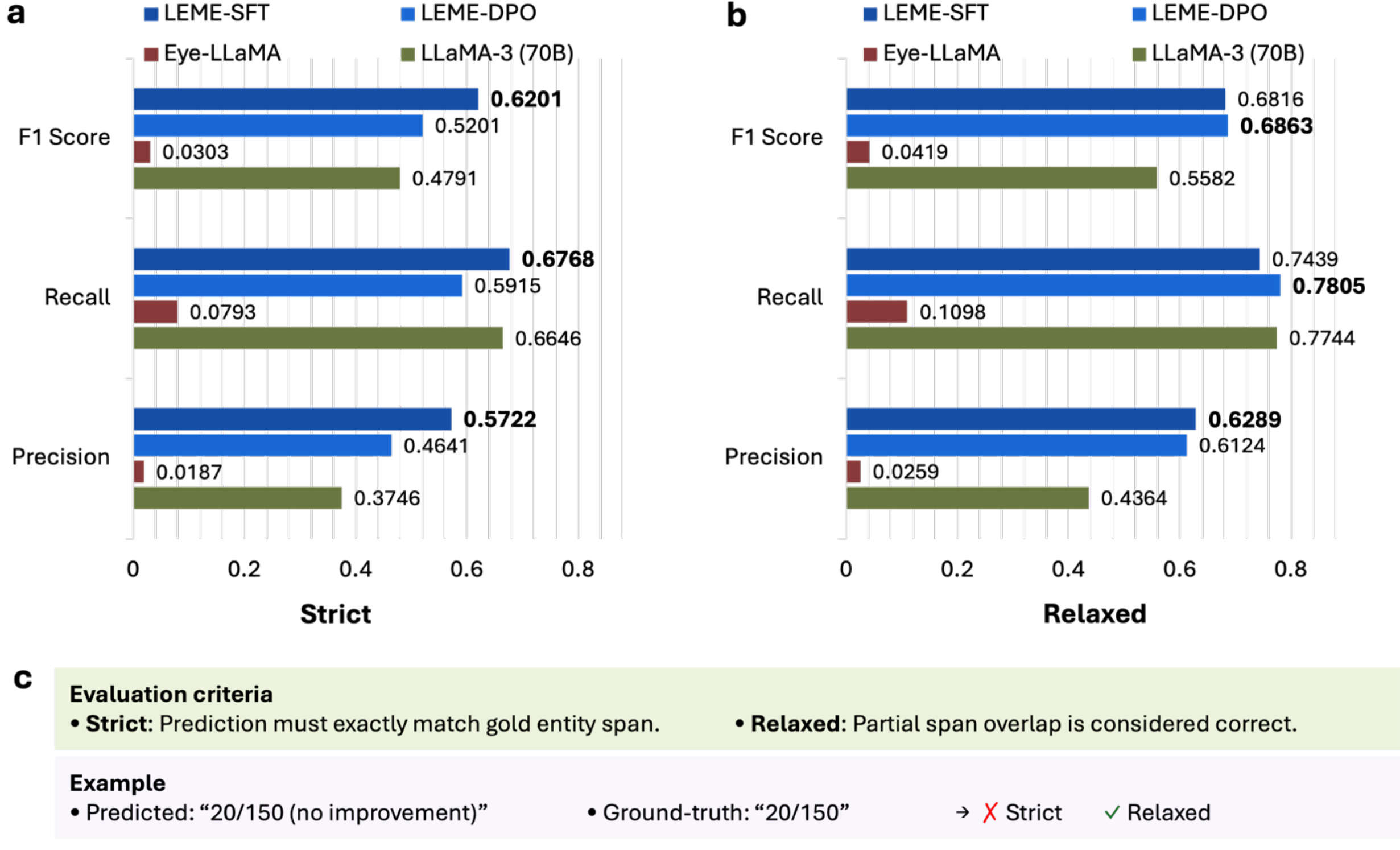

**Fig. 4 | Performance on the visual acuity extraction task under strict and relaxed evaluation criteria. a,** Results under the strict evaluation criterion (exact span match required). **b,** Results under the relaxed criterion (partial span overlap credited). **c,** Illustration of the evaluation criteria.

### Visual acuity extraction from clinical notes

In addition to evaluations based on publicly available data, we assessed LEME on a real-world clinical task using progress notes from Yale New Haven Hospital System: extracting visual acuity (VA) information. These values are often documented in inconsistent formats, making manual review labor-intensive and error-prone. Automated extraction can support downstream applications such as cohort identification, clinical decision support, and outcome tracking[44-46]. We manually annotated 300 notes with eight predefined VA entity types (see **Extended Data Table 3** for entity descriptions and dataset statistics). We randomly split the dataset into 200 notes for zero-shot testing and 100 notes for fine-tuning, allowing us to assess LEME's performance in both zero-shot and fine-tuned settings for VA extraction. We evaluated model performance using precision, recall, and F1 score, under both strict and relaxed matching criteria. The strict metric required an exact match between predicted and ground-truth entities. The relaxed metric awarded partial credit when predicted entities either contain or are contained by the ground-truth entities. Both strict and relaxed matches are commonly used for information extraction evaluations[52,53].

**Zero-shot performance** We report zero-shot performance under both strict (**Fig. 4a**) and relaxed (**Fig. 4b**) evaluation settings. Under the strict setting, LEME-SFT achieved the highest F1 score of 0.6201, followed by LEME-DPO (0.5201), LLaMA-3 70B (0.4791), and Eye-LLaMA (0.0303). Furthermore, LEME-SFT (0.5722) and LEME-DPO (0.4641) both achieved higher precision, than LLaMA-3 (0.3746) and Eye-LLaMA (0.0187). This pattern persisted under the relaxed criterion. LEME-SFT (0.6816) and LEME-DPO (0.6883) also achieved the highest F1 scores, substantially surpassing LLaMA-3 (0.5582) and Eye-LLaMA (0.0419). Precision remained strong in this setting as well: LEME-SFT and LEME-DPO recorded 0.6289 and 0.6124, outperforming LLaMA-3 (0.4364) and Eye-LLaMA (0.0259).

**Low-supervision adaptation** We further fine-tuned LEME for visual acuity extraction using 100 annotated patient notes containing 86 entity mentions. This setup reflects a typical low-supervision scenario in ophthalmology, where labeled data are often scarce. We applied LoRA, allowing rapid adaptation with minimal data[54]. **Extended Data Table 4** shows that both LEME-SFT and LEME-DPO maintained strong performance, achieving F1 scores of 0.7692 and 0.7491, respectively, under the strict criterion; and 0.7808 and 0.7960, under the relaxed criterion. LLaMA-3 performed worse, with scores of 0.7344 (strict) and 0.7516 (relaxed).

**Qualitative analysis** LEME-DPO often extracted not only the visual acuity mention itself but also additional surrounding context that conveyed clinically useful information. While these outputs were accurate in meaning, and in some cases, provided even more relevant detail than the gold annotation. However, they were penalized under the strict evaluation due to span mismatches. For instance, when the model predicted "20/150 (no improvement with refraction)" but the gold label included only "20/150," the result was marked incorrect under the strict criterion. However, this broader span sensitivity proved advantageous under the relaxed evaluation, where partial overlaps between predicted and ground-truth entities were credited.

A common error mode in LLaMA-3 involved overprediction in instances where multiple entity types shared the same surface form. For example, when "20/20 -1" corresponded to both right and left eye distance visual acuities without spectacle correction, LLaMA-3 frequently over-predicted four entity types, wrongly including both right and left eye distance acuities with spectacle correction. In contrast, both LEME models correctly identified only the two appropriate types.

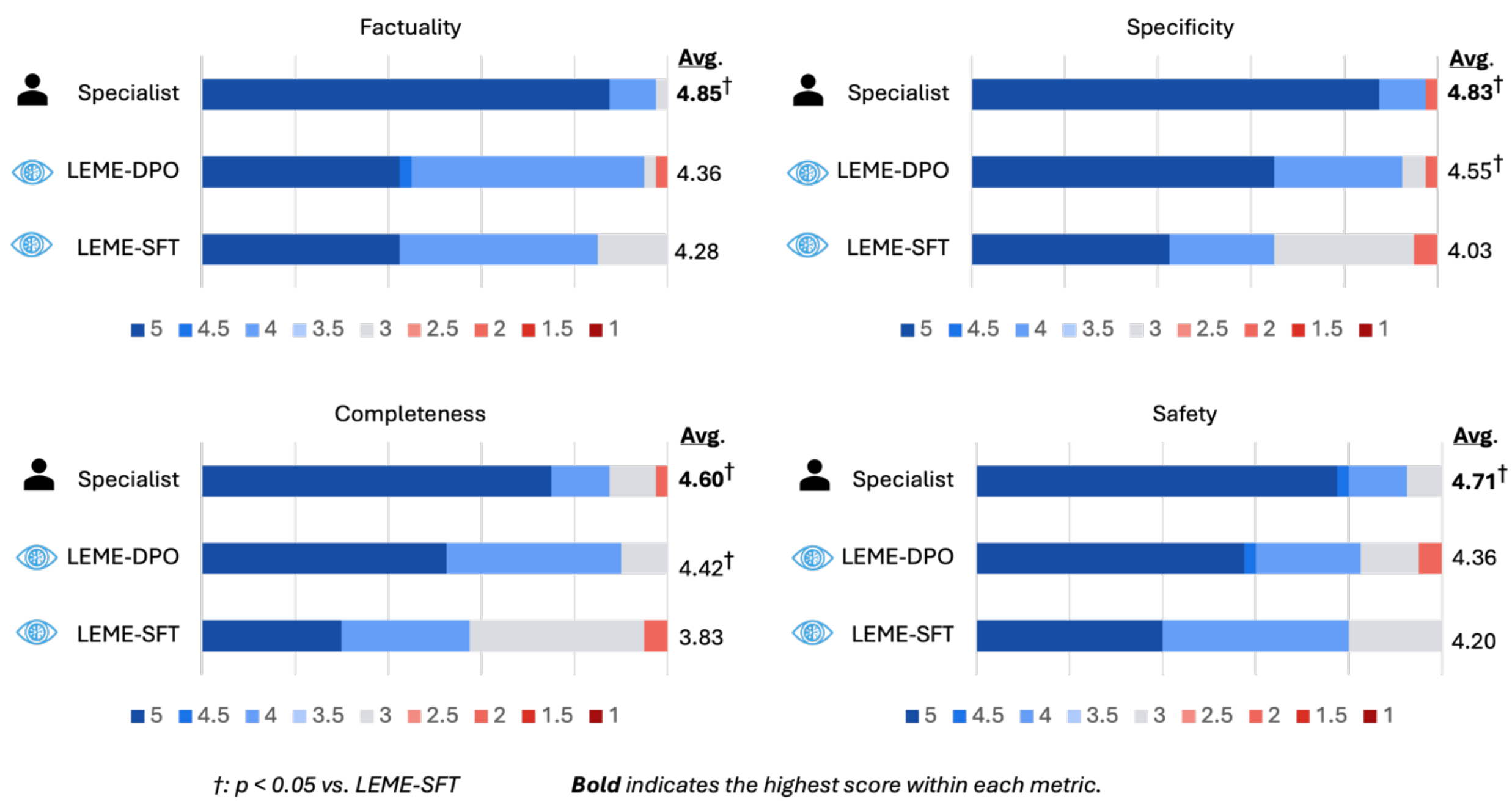

**Fig. 5 | Expert evaluation of end-to-end assessment and treatment planning on 40 patient cases**. Clinician-authored documentation (Specialist) and model-generated outputs (LEME-SFT, LEME-DPO) were evaluated on 40 patient cases. Average scores are shown on the right. LEME-DPO significantly outperformed LEME-SFT in specificity and completeness ($p < 0.05$; one-sided permutation tests), with performance approaching that of attending clinicians. Bold values indicate the highest score within each metric.

While this type of overprediction was common in baseline models, we observed no strong pattern of performance variation across specific entity types (see **Extended Data Table 5** for a breakdown).

## End-to-end assessment and treatment planning on real patient cases

We conducted a pilot evaluation on assessment and treatment planning using progress notes and imaging summaries from real patient visits. This setup mirrors routine clinical workflows, where clinicians integrate information from disparate sources to formulate diagnostic and therapeutic decisions[55,56]. We evaluated the two best-performing model variants from earlier evaluations, LEME-SFT and LEME-DPO, against the original documentation written by attending clinicians. Baseline visits from 40 patient cases were randomly selected, each diagnosed with diabetic retinopathy, age-related macular degeneration, or glaucoma. We manually removed the original assessment and treatment plans written by attending clinicians from the progress notes. The masked progress notes and associated imaging reports of each visit were then provided to LLMs to generate automated assessment and treatment plans. All the responses (LLM responses and the original assessment and treatment plans by attending clinicians) are anonymized and randomly shuffled. A senior ophthalmologist with over 30 years of clinical experience independently evaluated the anonymized responses using a five-point scale across four dimensions: factuality, specificity, completeness, and safety.

As shown in **Fig. 5**, the clinician-authored documentation received the higher average scores across all evaluation criteria: factuality (4.85), specificity (4.83), completeness (4.60), and safety (4.71). These high performances were expected as they were documented by attending-level clinicians. LEME-DPO achieved scores of 4.36 for factuality, 4.55 for specificity, 4.42 for completeness, and 4.36 for safety, lower than the experts' documentation yet not substantially inferior. Given the complexity of the task and the high bar set by clinician-authored notes, these results underscore the potential of LEME-DPO to support nuanced clinical documentation. LEME-SFT, by comparison, received 4.28 for factuality, 4.03 for specificity, 3.83 for completeness, and 4.20 for safety. Overall, LEME-DPO outperformed LEME-SFT in both specificity (4.55 vs. 4.03) and completeness (4.42 vs. 3.83), all $p < 0.008$.

# Discussion

We introduce LEME, a suite of state-of-the-art open-weight LLMs for ophthalmology. LEME addresses current key gaps in the development and evaluation of LLMs in ophthalmology. First, LEME leverages the most comprehensive, license-permissive ophthalmology corpus to date, curated from clinical guidelines, textbooks, and case reports for research use. Second, it employs a two-stage training framework that combines reasoning-enhanced supervised fine-tuning to strengthen domain-specific reasoning and instruction-following, with reinforcement learning to align model outputs with standards of factuality and informativeness. Third, LEME is systematically evaluated across clinically relevant benchmarks and downstream tasks using real-world clinical data, including a challenging end-to-end generation of assessment and treatment plan documentation. The data and training framework of LEME are designed for sustainability and can be readily adapted to future emerging LLM architectures for domain adaptation and fine-tuning. To foster future research and broader adoption, we publicly release the data, code, and model weights for non-commercial research use.

Evaluation across the five benchmarks consistently demonstrated that LEME significantly outperformed all seven baseline LLMs. On average, LEME achieved improvements of 3.32 and 4.60 percentage points over state-of-the-art proprietary models GPT-4o and o1-mini, respectively, and 8.79 points over LLaMA-3 70B, which served as the backbone architecture for LEME. For Eye-LLaMA, the only publicly available ophthalmology-specific LLM, evaluation was possible on three of the five benchmarks due to training data overlap in the others. Across these three benchmarks, LEME outperformed Eye-LLaMA by an average of 12.31 percentage points. In addition, prior studies have predominantly relied on multiple-choice QA benchmarks to evaluate LLMs in ophthalmology and more broadly in medicine[28,33-37,40,57]. Despite being straightforward and scalable, they are unrealistic in clinical practice—where clinicians rarely answer predefined-choice questions—and offer limited insights into downstream utility[26-28,31,32]. In contrast, our five benchmarks are open-ended and more clinically relevant, covering tasks such as answering patient questions about eye diseases, supporting patient-physician consultations, and planning treatments. To complement these evaluations, we also assessed LEME on the Basic and Clinical Science Course (BCSC) question set, a standard ophthalmology curriculum and the most used benchmark in prior LLM studies. With permission from the American Academy of Ophthalmology (AAO), we were authorized to evaluate up to 270 multiple-choice questions from the BCSC. LEME achieved an accuracy of 80.60%, outperforming all other open-weight LLMs. Detailed results are provided in **Extended Data Table 6**. A manual review by an attending ophthalmologist further confirmed LEME's stronger reasoning on BCSC questions: compared to LLaMA-3, LEME was preferred in 13 out of 25 cases, judged equivalent in 10, and outperformed in only 2 (see **Supplementary Information, section 1** for details).

For open-ended tasks, we used ROUGE-L and BERTScore as proxies for lexical and semantic similarity. While these metrics, along with our curated benchmarks, represent a meaningful improvement over prior knowledge-testing-based evaluations in ophthalmology, they do not fully capture clinical relevance[58,59]. In addition, these benchmark tasks were not based on real-world clinical notes. To address these gaps, we complemented them with three downstream evaluations that incorporated real-world clinical data and direct involvement of clinicians, enabling assessment of model performance in more authentic and practice-relevant settings. First, manual review of real-world patient Q&A pairs showed that LEME received higher ratings than LLaMA-3 across four key dimensions: factuality, specificity, completeness, and safety. This is consistent with prior findings that general-purpose LLMs underperform in specialized domains and benefit from domain-specific adaptation[60-62]. Second, in a zero-shot task extracting visual acuity from clinical notes, LEME significantly outperformed all baselines. Despite not being trained on this type of data, it showed strong generalizability to unseen clinical content. Fine-tuning on a small set of notes further improved performance, indicating adaptability to varied settings. Third, in generating assessments and treatment plans from patient visits, LEME-DPO showed performance approaching that of expert-written documentation, with clear improvements over LEME-SFT. This highlights the contribution of reinforcement learning in improving alignment with clinical expectations.

We further address the reproducibility and accessibility challenges observed in prior ophthalmology-specific models. Open-weight LLMs are critical for medical applications, as proprietary systems are often constrained by access restrictions and patient privacy concerns[23,24,63,64]. However, previous models present limitations that hinder broader utility. For example, Ophtha-LLaMA 2 was trained on proprietary hospital data, limiting external use[26]. Eye-LLaMA and EyeGPT rely on AAO content, which is not licensed for AI training[27,28]. Additionally, many existing models lack public checkpoints or do not disclose detailed training data, hindering reproducibility and independent evaluation[26,27,31,32]. In contrast, LEME is the first ophthalmology-specific LLM that is both openly available and designed to enable broad research use. It is trained exclusively on publicly available data with verified licenses. While the current release is restricted to research settings, the same training pipeline can readily accommodate commercial datasets for proprietary applications.

We also offer practical recommendations for optimal use of LEME based on our evaluation results. LEME is released in two versions, LEME-SFT and LEME-DPO, corresponding to the supervised fine-tuning and preference optimization stages of the training framework. LEME-SFT, trained exclusively with supervised fine-tuning, delivers concise and focused outputs. Its brevity is particularly advantageous for tasks such as summarization and information extraction. This could be especially valuable for clinicians who need to distill key medical history and critical clinical findings from the excessive imaging data and text information accumulated during follow-up visits. By streamlining these processes, LEME-SFT has the potential to significantly enhance working efficiency and facilitate smoother clinical workflows. On the other hand, LEME-DPO, enhanced through preference optimization, generates more comprehensive and detailed responses. This makes it better suited for tasks requiring in-depth clinical reasoning, such as performing differential diagnoses and formulating treatment plans. Beyond its practical applications, LEME-DPO can serve as a critical tool for cultivating junior clinicians' analytical and decision-making skills. Its interpretability and transparency grounded by rationality further offers reassurance for clinical decisions, representing a pivotal step for the active integration of LLMs into real-world clinical practice.

This study has several limitations. First, although we conducted expert-based clinical validation, our evaluations still did not cover the full spectrum of downstream clinical ophthalmology applications. As such, while our findings offer valuable insights, further validation is warranted, particularly in scenarios where longitudinal data is involved. Second, we used an LLM-argument pipeline guided by clinician-crafted instructions and prompts and filtered through consistency checks. While this approach improves scalability compared to full manual curation, its quality and cost-effectiveness (e.g., use of proprietary LLM) remain to be evaluated. Lastly, although we conducted three downstream evaluations beyond standard benchmarks, future prospective studies across multiple clinical centers are needed. Such studies will be critical for assessing the robustness of models across varying EHR formats, documentation practices, and patient populations, as well as its impact on clinical workflows and outcomes.

In conclusion, LEME is the first openly available, license-compliant LLM suite for ophthalmology. By combining a comprehensive domain corpus, a two-stage training framework, and rigorous multi-faceted evaluation, LEME consistently outperforms all evaluated baselines on clinically relevant tasks. Its dual versions support both concise information extraction and complex reasoning, enabling applications from streamlining documentation to diagnostic decision-support. These capabilities have the potential to enhance clinical efficiency, improve the quality and safety of patient care, and contribute to the education of future ophthalmologists. While broader, multi-center validation remains important, LEME provides a reproducible, adaptable foundation for the safe, effective, and scalable integration of LLMs into clinical ophthalmic practice.

# Methods

### Development of LEME

**Model architecture** The model architecture employed in this study follows a Transformer decoder-only design[65], as used in the LLaMA family of models. It consists of stacked decoder layers with multi-head self-attention and feed-forward networks, enabling autoregressive generation by attending to previously generated tokens. Positional embeddings are added to input token embeddings to preserve token order. This architecture excels at sequence modeling and text generation tasks due to its capacity to capture long-range dependencies. Although our experiments use this decoder-only backbone, the training and fine-tuning procedures described are agnostic to the underlying architecture, meaning other architectures, such as encoder-decoder or decoder-only models with similar capabilities, can be used without fundamental changes to the methodology.

**Ophthalmology corpora** The development of LEME followed a structured, multi-stage approach, starting with the careful selection of corpora based on open-access availability, license compatibility, authoritative sourcing, and the inclusion of practice-oriented clinical content. Three corpus types were curated: clinical guidelines, StatPearls (textbooks), and case reports. Detailed information for these corpora, including statistics, licenses and download links, are provided in **Extended Data Table 7**. As shown in **Fig. 6a**, First, nine reputable clinical guideline providers, including the National Institute for Health and Care Excellence (NICE), Centers for Disease Control and Prevention (CDC), World Health Organization (WHO), Canadian Medical Association (CMA), Cancer Care Ontario (CCO), Strategy for Patient-Oriented Research (SPOR), International Committee of the Red Cross (ICRC), PubMed, and WikiDoc, using the Meditron guidelines corpus. These documents were segmented into 1,000-word snippets and classified using GPT-4o to identify 3,435 ophthalmology-relevant entries (see **Supplementary Table 2** for the prompt). Next, we incorporated 147,985, preprocessed from the MIRAGE project[41] (publicly available benchmarks curated for retrieval-augmented generation with LLMs) and applied the same GPT-4o classification to extract 12,919 ophthalmology-focused texts. Finally, we filtered the PMC-Patients-V2 dataset (a dataset of open-access patient case reports)[42], comprising 250,294 open-access case reports from the 2024 PMC baseline, by matching a list of 105 ophthalmology journals, resulting in 5,745 ophthalmology-specific case reports. Those processed corpora were used to generate weak labelled instances for training below.

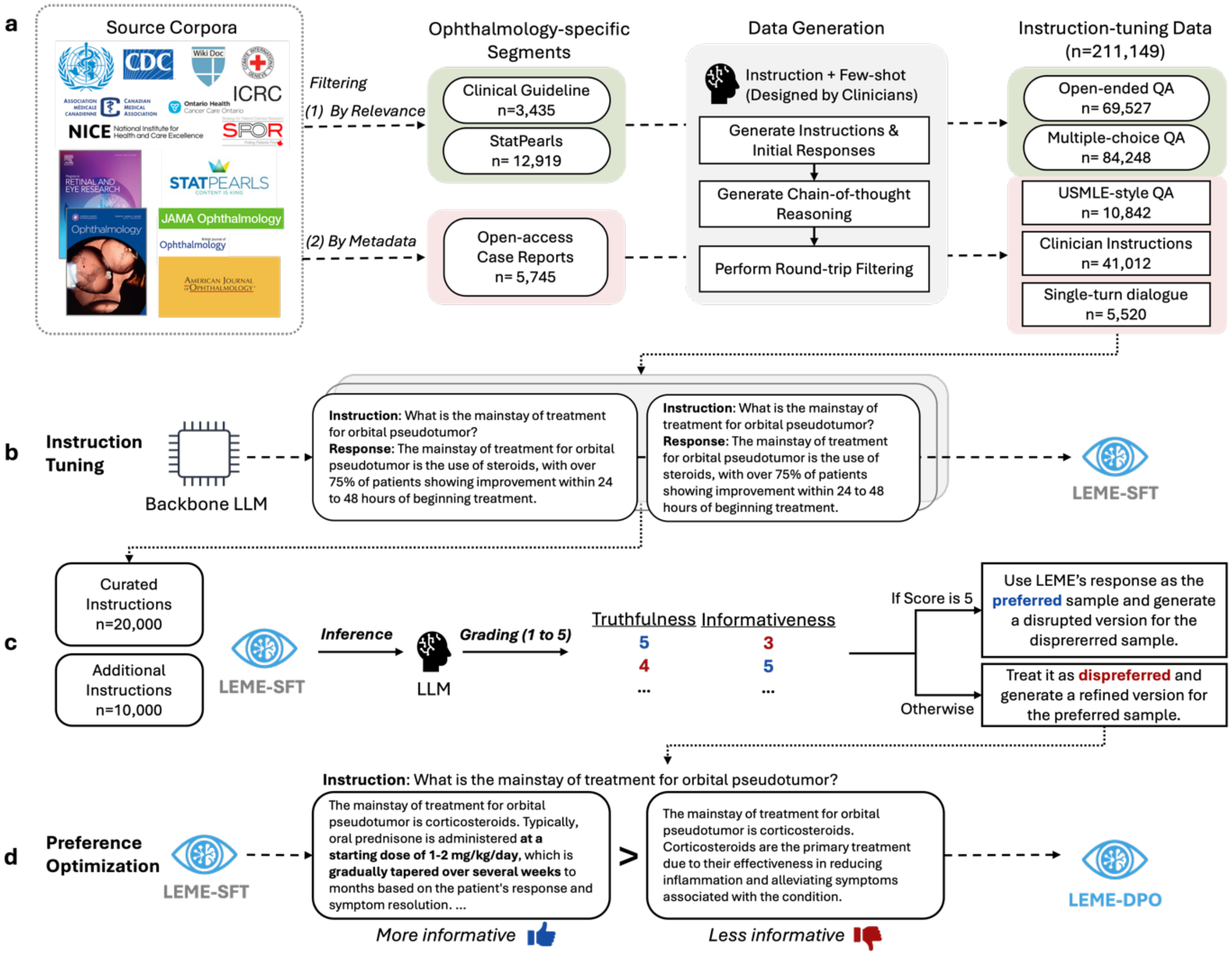

**Fig. 6 | Development pipeline for LEME via instruction-tuning and preference optimization. a,** Instruction-tuning data construction from three ophthalmology-specific sources: clinical guidelines, StatPearls, and real-world case reports. Relevant texts were filtered based on content and metadata, followed by task-specific prompt generation, yielding 211,149 instruction samples. **b,** Instruction-tuning: the curated instructions were used to fine-tune the base LLM with step-by-step reasoning and ophthalmology-focused tasks, resulting in LEME-SFT. **c,** Construction of preference-labeled data: 29,747 preferred/dispreferred pairs were generated by evaluating responses for truthfulness and informativeness, using outputs from LEME and samples from BioInstruct. **d,** Preference optimization: LEME-DPO was trained on the preference data to improve response quality, with supervision provided via LLM-based evaluations.

**Instruction-tuning data** Using the ophthalmology-specific corpora, we generated a total of 211,149 instructions, with formats tailored to the distinct characteristics of each corpus type. These instructions were constructed using an LLM-augmented pipeline, guided by clinician-drafted instructions and prompts, and filtered using a roundtrip consistency strategy to select high-quality instructions[66]. This approach enables large-scale generation of weakly labeled training data and compatible with different LLM backbones. We used GPT-4o as a representative LLM to implement the pipeline. **Fig. 6a** outlines the overall process.

For clinical guidelines and StatPearls, sources of foundational medical knowledge, we focused on QA-based instruction formats. We generated both open-ended and multiple-choice QA pairs by prompting GPT-4o with five manually crafted few-shot examples by clinicians. GPT-4o then generated answers for each question. For open-ended QA, the initial long-form responses were reviewed and refined for quality. For multiple-choice QA, step-by-step reasoning traces were generated to support the selected answers. To ensure quality, we adopted a roundtrip consistency approach, retaining only instances where the model correctly answered its own question. The reasoning traces were preserved for use in LEME training. This process yielded 69,527 open-ended QA pairs and 84,248 multiple-choice QA pairs. Please refer to **Supplementary Tables 3 and 4** for the generation of open-ended and multiple-choice QA pairs, respectively. See **Supplementary Table 5** for the open-ended QA refinement prompt and **Supplementary Table 6** for the multiple-choice reasoning prompt.

In contrast, the case report corpus, rich in-patient narratives and clinical scenarios, informed the design of three task types: USMLE-style QA, clinician-style information extraction prompts, and single-turn patient-clinician dialogues. For USMLE-style QA, we used few-shot prompts with examples and applied roundtrip consistency filtering (see **Supplementary Table 7** for the prompt). The few-shot examples were randomly sampled from a pool of 455 eye health-related questions extracted from the MedQA training set[67], with three examples selected per prompt. For clinician-style tasks, GPT-4o was given 15 seed instructions drafted by clinicians, which it could reuse or adapt to simulate tasks such as summarization and information extraction. Refer to **Supplementary Table 8** for the full list of seed instructions and **Supplementary Table 9** for the prompt. In the dialogue task, GPT-4o generated realistic single-turn interactions, pairing patient queries with clinician responses (see **Supplementary Table 10** for the prompt). This resulted in 10,842 USMLE-style QA items, 41,012 clinician prompt-based instructions, and 5,520 single-turn dialogue examples. Importantly, these LLM-augmented instructions were used only for model training as described below, while all evaluations were conducted using manually curated datasets.

**Instruction tuning** We applied instruction tuning to adapt the language model to follow diverse natural language instructions relevant to ophthalmology tasks (**Fig. 6b**). Specifically, the model is trained to generate accurate responses conditioned on input prompts comprising task instructions and contextual information. This training optimizes the standard autoregressive cross-entropy loss, maximizing the likelihood of the output tokens $y = (y_1, \dots, y_T)$ given the input sequence $x$:

$$L(p_\theta) = -\sum_{t=1}^{T} \log\ p_\theta(y_t | y_{<t}, x),$$

where $T$ is the length of the output sequence, and $p_\theta(\cdot \mid \cdot)$ denotes the model's predicted token distribution at step $t$, conditioned on the previous tokens and input. This objective enables the model to generalize across a wide range of task formats, including open-ended question answering, multiple-choice QA, summarization, and information extraction, by learning to map natural language instructions to appropriate outputs. Instruction tuning thus improves the model's ability to perform complex, domain-specific clinical tasks by grounding responses in both task context and domain knowledge. In this study, we leverage richly curated instruction–response pairs derived from ophthalmology-specific corpora to guide the model toward domain-specialized reasoning and language generation.

**Preference-labeled data** While instruction tuning enables models to follow task-specific prompts, it does not explicitly optimize for key quality attributes such as factuality and safety[30,68-70]. To address this, we introduced a preference optimization stage that further trains the instruction-tuned model to distinguish and prioritize high-quality responses over suboptimal ones across ophthalmology and broader medical tasks. We created a preference dataset of 29,747 pairs, each consisting of a preferred and a dispreferred response, using a similar LLM-augmented approach (**Fig. 6c**). This dataset was built by sampling 20,000 examples from our instruction-tuning data and 10,000 from BioInstruct[71] (https://huggingface.co/datasets/bio-nlp-umass/bioinstruct), a diverse biomedical instruction dataset covering tasks such as information extraction, note summarization, simplification, and diagnosis. Although BioInstruct is not ophthalmology-specific, its inclusion aimed to improve the model's ability to generalize across varied tasks. For 30K instruction inputs, responses were generated using the LEME-SFT checkpoint and evaluated by GPT-4o for truthfulness and informativeness (see **Supplementary Tables 11 and 12** for the prompts). Responses scoring 5 on truthfulness were labeled preferred, with corresponding dispreferred examples created by partially modifying the original to reduce accuracy while maintaining structure (**Supplementary Table 13**). Conversely, responses scoring below 5 were marked dispreferred, and an improved version was generated as preferred (**Supplementary Table 14**). Subsequently, the same procedure was applied for informativeness evaluation, considering only responses with a truthfulness score of 5 (**Supplementary Tables 15 and 16**). After filtering out samples with generation errors, the final dataset included 29,747 preference pairs.

**Preference optimization** To further refine response quality, we applied preference optimization using Direct Preference Optimization (DPO) on a dataset of 29,747 labeled preference pairs (**Fig. 6d**). Each pair consists of a preferred and a dispreferred model-generated response, annotated based on criteria such as truthfulness and informativeness as assessed by an LLM. DPO optimizes the model to assign higher likelihoods to preferred responses over dispreferred ones without requiring explicit reward modeling. Formally, given an input $x$ and two responses $y^+$ (preferred) and $y^-$ (dispreferred), DPO maximizes the following objective:

$$L(p_\theta; p_{\text{ref}}) = \log \sigma \left( \beta \log \frac{p_\theta(y^+|x)}{p_{\text{ref}}(y^+|x)} - \beta \log \frac{p_\theta(y^-|x)}{p_{\text{ref}}(y^-|x)} \right),$$

where $\sigma$ is the sigmoid function, $\beta$ is a hyperparameter for temperature scaling, $p_\theta$ is the model being trained, and $p_{\text{ref}}$ is a reference model (i.e., the model after instruction tuning). This objective encourages the model to increase the probability gap between preferred and dispreferred responses conditioned on the same input. By training on this preference-labeled dataset, the model learns to generate outputs that better satisfy nuanced quality criteria, improving clinical relevance and linguistic accuracy beyond instruction tuning alone. This stage produced the LEME-DPO model variant, which demonstrates enhanced performance on downstream ophthalmology tasks.

**Implementation details** Our model was initialized with the LLaMA-3-70B-Instruct model (https://huggingface.co/meta-llama/Meta-Llama-3-70B-Instruct) and trained using the LLaMA-Factory library[72] on eight A100 GPUs (80GB each) with DeepSpeed ZeRO-3 and CPU offloading enabled. The LEME-SFT model was trained for 3 epochs with a batch size of 64 and a learning rate of $5.0e^{-6}$. Subsequently, LEME-DPO was fine-tuned from the LEME-SFT checkpoint using the preference dataset for 1 additional epoch with a batch size of 32 and a learning rate of $5.0e^{-7}$.

## Benchmark evaluation on five open-ended tasks

**Evaluated models** We selected seven LLMs for comparison, encompassing a range of model types and domains. These include GPT-4o[20], OpenAI's latest state-of-the-art general-purpose model; and o1-mini[21], a smaller variant of OpenAI's reasoning-focused o1 model, chosen here due to API cost considerations. As representatives of open-weight general-domain models, we included LLaMA-2 (7B and 70B)[48] and LLaMA-3 70B[47], using their publicly available instruction-tuned versions. Meditron 70B[40] was selected as a medical-domain model representative, developed by continuing the pretraining of LLaMA-2 70B on biomedical corpora such as PubMed, PMC, and clinical guidelines. Lastly, Eye-LLaMA (7B)[28], as mentioned, currently the only open weight model specifically tailored for ophthalmology using LLaMA-2 7B as the backbone. During inference, all models were evaluated in a zero-shot setting using standardized instruction templates (see **Supplementary Information, section 3**). An exception was made for Meditron: its original study only provided the pre-trained version, which did not follow instructions under the zero-shot setting; therefore, we evaluated it using a few-shot approach instead. The temperature was set to 0 to minimize response variance.

**Evaluation methods** For open-ended tasks, we used ROUGE-L[73] and BERTScore[49], two widely adopted metrics for natural language generation. ROUGE-L measures the longest common subsequence between the generated and reference texts, capturing lexical overlap and favoring fluent responses that preserve the reference structure. In contrast, BERTScore leverages contextual embeddings from pre-trained language models to assess semantic similarity, providing a more nuanced evaluation of meaning alignment even when surface forms differ. Together, these metrics offer complementary perspectives on both surface-level fidelity and deeper semantic relevance. For statistical comparison, we performed 10,000 rounds of bootstrap resampling using the same number of examples as the original dataset and applied Holm-Bonferroni correction to adjust for multiple comparisons.

**Patient QA** We utilized long-form question–answer pairs sourced from the American Academy of Ophthalmology's (AAO) Ask an Ophthalmologist forum, where certified ophthalmologists respond to patient health inquiries. The AAO has granted permission to use these 76 instances for evaluation purposes. Questions were randomly sampled from various subtopics, including Retina, Glaucoma, Cataracts, Dry Eye, and Uveitis, to ensure broad topical coverage. This dataset enables evaluation of model performance on expert-generated, clinically relevant patient queries.

**Public eye health QA** Derived from educational content provided by the National Eye Institute (NEI) of the National Institutes of Health (NIH), this dataset covers foundational knowledge of common eye conditions. Each condition is organized into question-based sections—such as definitions, symptoms, causes, and treatments—each posed as a user-friendly question (e.g., What causes cataracts?) followed by accessible explanations. We framed these as 281 long-form QA pairs to evaluate models' ability to generate accurate, comprehensible medical information suitable for patient education.

**Patient consultation** The dataset originates from iCliniq, an online health consultation platform. We used the data released in prior research, which comprises approximately 7.3K questions[74]. From this, we used GPT-4o to identify and extract 303 questions specifically related to eye health, followed by manual review for relevancy. The data reflects diverse, informal, user-generated question styles common in online medical communities, offering a valuable testbed for assessing model robustness to varied input quality and informal language.

**Treatment planning** This dataset consists of 245 clinical cases sourced from EyeRounds, a resource provided by the Department of Ophthalmology at the University of Iowa. Each case includes an initial presentation detailing patient history, ocular examination findings, and diagnosis, which serves as the input. The associated treatment summary is used as the ground truth output, allowing evaluation of a model's capability to generate appropriate clinical treatment plans from patient case information.

**Abstract-to-title generation** Comprising 500 abstracts from ophthalmology research articles published in peer-reviewed journals, this dataset is designed to generate corresponding article titles based on abstract content. This task measures a model's ability to accurately distill key scientific information and produce concise, informative titles, thereby testing summarization and scientific comprehension skills. Additionally, it highlights the model's potential utility as a research assistance tool.

**Question answering on patient symptom and concern queries**

**Data preparation** We initially collected the 76 question-expert answer pairs licensed by the American Academy of Ophthalmology for model evaluation purposes as described in the **Patient QA** section. The original set included knowledge-based questions. To exclude them, we manually selected 41 examples that explicitly referenced patient-specific information such as age, gender, symptoms, or chief complaints. Of these, two examples, along with several general knowledge questions, were used to align evaluation criteria between the two evaluators, and the remaining 39 were used for the final evaluation.

**Evaluation methods** We conducted human evaluation across four dimensions: (1) Factuality, which assesses whether responses are consistent with expert ophthalmologic knowledge and established clinical guidelines; point deductions are applied for incorrect information or unsupported speculation; (2) Specificity, which evaluates the level of detail; responses should avoid vague or generic advice and instead provide concrete, actionable guidance; (3) Completeness, which checks whether all essential information required to address the question is included; optional details may enhance clarity but are not mandatory; and (4) Safety, which ensures that responses do not contain or omit information that could lead to patient harm; potentially dangerous advice or the absence of a necessary referral to a healthcare professional are penalized. Each response begins with a score of 5. Minor issues incur a 1-point deduction, major issues 2 points. Partial deductions (e.g., 0.5 or 1.5) are allowed, but the maximum deduction per criterion is capped at 2 points. Final scores range from 1 to 5. Further details are provided in **Supplementary Information, section 4**.

Because the tasks demand nuanced clinical judgment, we recruited experienced ophthalmologists to rate each model response. A pilot round showed limited agreement among clinicians, mirroring real practice in which definitive answers are uncommon and expert interpretations naturally diverge. Rather than imposing artificial consensus, we averaged the individual scores. This approach preserves the diversity of expert opinions while providing a clear, practical metric for comparison. To reduce bias, all responses were anonymized, and the order of model outputs was randomized for each sample, preventing raters from identifying the source model or whether the response was generated by a human.

All p-values were obtained from pairwise permutation tests conducted with 10,000 iterations each. For every comparison, the test statistic (e.g., performance difference between two models) was computed under the null hypothesis by randomly shuffling the model labels and recalculating the metric across resampled datasets. The resulting p-values were then adjusted for multiple comparisons using the Benjamini–Hochberg procedure, with statistical significance determined at a threshold of $p < 0.05$ throughout.

### Visual acuity extraction from clinical notes

**Data preparation** The evaluations involving clinical data in this study is approved by the Yale Institutional Review Board (IRB) under protocol number 2000035967. We constructed a dataset of 300 progress notes from the Yale New Haven Hospital System, authored by approximately 40 different healthcare providers. We selected 250 notes that matched the regular expression pattern r”(?i)(visual\s+acuity)”, suggesting the presence of visual acuity (VA) mentions, and 50 notes that did not match this pattern, and were thus likely to lack any VA-related content. This setup enabled us to evaluate model performance in both VA-present and VA-absent scenarios. All 300 notes were manually annotated by a medical student using a predefined schema of eight clinical entity types (**Extended Data Table 3**), following standardized guidelines adapted from prior work[75].

### End-to-end assessment and treatment planning on real patient cases

**Data preparation** We began with a large pool of progress notes from the Yale New Haven Hospital System. We first identified notes that explicitly included assessment and treatment planning sections, and then filtered for patients diagnosed with diabetic retinopathy (DR), age-related macular degeneration (AMD), or glaucoma. For each selected visit, we also collected one or more imaging summaries (e.g., OCT, fundus photography) recorded within ±48 hours of the corresponding progress note to ensure temporal alignment. Subsequently, we manually removed the assessment and treatment sections from each progress note. We then carefully reviewed the remaining content, including both the unmasked content and associated imaging summaries, to confirm that no residual diagnostic or management information was present elsewhere. In parallel, we reviewed the original assessment and treatment plans (used as reference responses) to ensure they focused exclusively on ocular findings, excluding unrelated systemic issues such as headache or dizziness. After applying all selection and filtering criteria, we used 40 visits for the main evaluation. The final dataset included documentation authored by five different clinicians, capturing variability in writing styles and clinical reasoning.

**Evaluation methods** The evaluation followed the same criteria described in the "Question answering on patient symptom and concern queries" subsection of the Methods section. A single senior ophthalmologist with over 30 years of clinical experience rated each response along four dimensions: factuality, specificity, completeness, and safety. Detailed descriptions of these criteria are provided in **Supplementary Information, section 4**. To assess statistical significance, we performed pairwise permutation tests with 10,000 iterations, using the observed difference in mean scores as the test statistic. All p-values were corrected for multiple comparisons using the Benjamini–Hochberg procedure, with a significance threshold of $p < 0.05$.

## Author Contributions

H.K., A.G., Y.C.T., and Q.C. contributed to conception of the study and study design. H.K., X.A., A.G., Q.X., M.B.S., S.A., and Y.Y. contributed to acquisition of the data. R.S., and X.A. contributed to clinical data collection. H.K., X.A., and K.P. contributed to statistical analysis of data. H.K., X.A., and A.G. contributed to model development. X.A., S. Srinivasan., and A.G. contributed to analysis and interpretation of the data. H.K., X.A., Q.N.N., S.Y.W., and Q.C. contributed to the design and formulation of downstream tasks. S.A., G.D.Y., M.Z., D.Z.C., S. Sarrafpour, J.L., and L.V.D.P. contributed to qualitative analysis and manual evaluation. W.J.Z. contributed to computing resources. H.K. and S. Srinivasan contributed to the creation of visual materials, including figures and tables. H.K., X.A., S. Srinivasan, A.G., Q.X., K.P., K.Z., J.H., E.L., P.W., L.D.P., L.O.O., H.X., D.L., W.J.Z., R.A.A., Y.C.T., and Q.C. drafted the manuscript. Y.C.T. and Q.C. supervised the study, including research planning and execution. All authors read and approved the final version of the manuscript.

## Data/Code Availability

Data and code will be made available upon acceptance.

## Funding

Dr Yih-Chung Tham is supported by the National Medical Research Council of Singapore (NMRC/MOH/ HCSAINV21nov-0001). Dr. Qingyu Chen is supported by the National Institutes of Health (National Library of Medicine, 4R00LM014024 and 1R01LM014604). This work is also partly supported by the National Institutes of Health through grants 1UM1TR004906-01, and the Cancer Prevention and Research Institute of Texas through grant RP170668 (Dr W. Jim Zheng). These funders had no role in study design, data collection, data analysis, decision to publish, or preparation of the manuscript.

## Consent to Participate Declaration

Not applicable.

## Competing Interests

The authors declare no competing interests.

**Extended Data Table 1 |** Example data samples illustrating each ophthalmology-specific benchmark task.

| Task | Example Data | |
|---|---|---|
| Patient QA | *Description* | The task is to provide an appropriate and professional response to a clinical inquiry submitted by an individual seeking advice from an Ophthalmologist. |
| | *Input* | I had cataract surgery three weeks ago. My vision is great, however, I continue to have discomfort in this eye. It feels like I have a contact lens in that eye. I feel and see the edges of my lens. It also feels like the lens is floating on a liquid substance. Can you tell me if this is a normal part of the cataract surgery three weeks out? |
| | *Ground truth* | *The ground truth is not shown here, as it requires permission from the American Academy of Ophthalmology.* |
| Public Eye Health QA | *Description* | The task is to provide an appropriate and professional response to a clinical inquiry submitted by an individual seeking advice from an Ophthalmologist. |
| | *Input* | Am I at risk for AMD? |
| | *Ground truth* | Your risk for AMD increases as you get older. People age 55 and older are more likely to have AMD. The risk for AMD is also higher for people who:<br>Have a family history of AMD<br>Are Caucasian<br>Smoke<br>If you're at risk for AMD because of your age, family history, or other factors, it's important to get regular eye exams. Ask your doctor how often you need to get eye exams. Early AMD doesn't have any symptoms, so don't wait for your vision to change! |
| Patient Consultation | *Description* | The task is to provide an appropriate and professional response to a clinical inquiry submitted by an individual seeking advice from an Ophthalmologist. |
| | *Input* | Hello doctor, I recently have noticed a slight blurriness in my right eye. It is only noticeable when I close my left eye, and so I am not sure how long it has taken for this to develop. I am going to see an optometrist and doctor early next week but am currently on the road for work. My main question is that when I look out of the corner of my right eye the blurriness seems to subside and I can see usually. I am wondering if this is a good sign. I have been dealing with severe anxiety the past couple months, and I am worried that might have raised my blood pressure enough to hurt my eyesight. Eyesight when looking with both eyes seems to be normal. I have also noticed floaters in my eyes. Kindly help. |
| | *Ground truth* | Hello. This kind of a picture of eyesight becoming better on looking at specific directions are related to astigmatism. It is a very prevalent and correctable condition. There is no need to be anxious about it. Different reasons cause floaters. A dilated retinal examination will be able to help us get to the cause. Most of the conditions causing floaters are benign (not severe), but a few are dangerous, but all are treatable. So, I recommend that you visit your optometrist and in case you need any further information or help with refractive surgery, please schedule a consultation. I hope this helps you. |
| Treatment Planning | *Description* | The task is to provide a brief summary of the appropriate treatments for the patient described in the note. |
| | *Input* | # INITIAL PRESENTATAION<br><br>History of Present Illness:<br>An 18-year-old male presented to clinic for regular monitoring of ocular manifestations of methionine synthase deficiency type homocystinuria, specifically cobalamin deficiency of the Cb1G subtype. He complained of some photophobia and intermittent inward turning of his eyes. […]<br>Past Ocular History:<br>Horizontal pendular nystagmus<br>Esotropia status post bilateral medial rectus recessions of 5.5 mm<br>Mild amblyopia OS, treated with atropine drops in the right eye (OD)<br>Past Medical History:<br>[…]<br>Medications:<br>[…]<br>Family History:<br>No family history of retinal disease<br>Review of Systems:<br>Negative except as documented in the history of present illness<br><br># OCULAR EXAMINATION<br><br>Visual Acuity with correction (Snellen):<br>[…] |
| | *Ground truth* | Vitamin supplementation and dietary restriction specific to the type of defect causing homocystinuria |
| | *Description* | The task is to generate the most appropriate paper title based on the given ophthalmology-related abstract. |
| Abstract-to-Title Generation | *Input* | Endothelin-1 (ET-1) is a potent vasoconstrictor peptide produced by vascular endothelial cells. ET-1 may have a role in the pathogenesis of various vascular diseases. There are reports in the literature that ET-1 plasma levels are raised in normal-tension glaucoma (NTG) patients. ET-1 concentration, plasma renin activity, and 24-h blood pressure were measured in 21 high-tension glaucoma (HTG) patients, 19 NTG patients, and 20 non-glaucomatous controls in supine and upright positions. ET-1 plasma levels tended to be higher in NTG patients (3.2 +/- 2.2 pg/ml) than in HTG patients (2.2 +/- 0.6 pg/ml) and controls (2.6 +/- 0.7 pg/ml). The differences, however, were not statistically significant. The individual scatter was significantly greater in the NTG group, indicating that our NTG patients are a heterogeneous population. The physiological increase in ET-1 plasma level after changing from the supine to the upright position was absent in NTG patients. Plasma renin activities tended to be lower in NTG patients (1.2 +/- 1.2 ng/ml/h) than in HTG patients (1.3 +/- 0.8 ng/ml/h) and controls (2.0 +/- 1.7 ng/ml/h). This may explain why NTG patients had relatively low blood pressure despite high ET-1 levels. Our data support the hypothesis that vascular dysfunction may be involved in the pathogenesis of optic nerve damage in normal-tension glaucoma. |
| | *Ground truth* | Endothelin-1 plasma levels in normal-tension glaucoma: abnormal response to postural changes |

**Extended Data Table 2 |** Evaluation of benchmark performance on five open-ended tasks.

| Task | Ours | | Publicly Available Models | | | | | Proprietary Models | |
|---|---|---|---|---|---|---|---|---|---|
| | LEME-SFT (70B) | LEME-DPO (70B) | EYE-LLaMA (7B) | MediTron (70B) | LLaMA-2 (7B) | LLaMA-2 (70B) | LLaMA-3 (70B) | GPT-4o | o1-mini |
| **ROUGE-L** | | | | | | | | | |
| Patient QA | **0.1953** (0.1941,0.1965) | 0.1738 (0.1729,0.1747) | N/A | 0.0513 (0.0506,0.0521) | 0.1295 (0.1286,0.1303) | 0.1335 (0.1326,0.1344) | 0.1327 (0.1318,0.1337) | 0.1535 (0.1526,0.1544) | 0.1298 (0.1290,0.1305) |
| Public Eye Health QA | **0.1935** (0.1927,0.1943) | 0.1708 (0.1702,0.1714) | 0.1426 (0.1421,0.1432) | 0.0502 (0.0499,0.0505) | 0.1335 (0.1330,0.1340) | 0.1408 (0.1402,0.1414) | 0.1375 (0.1370,0.1380) | 0.1512 (0.1506,0.1518) | 0.1219 (0.1213,0.1224) |
| Patient Consultation | **0.1478** (0.1474,0.1481) | 0.1296 (0.1293,0.1300) | N/A | 0.0385 (0.0382,0.0387) | 0.1105 (0.1102,0.1109) | 0.1135 (0.1132,0.1139) | 0.1102 (0.1098,0.1105) | 0.1153 (0.1150,0.1157) | 0.1056 (0.1052,0.1059) |
| Treatment Planning | **0.1017** (0.1010,0.1024) | 0.0966 (0.0961,0.0971) | 0.0655 (0.0651,0.0659) | 0.0335 (0.0332,0.0338) | 0.0613 (0.0608,0.0617) | 0.0623 (0.0619,0.0627) | 0.0819 (0.0815,0.0824) | 0.0779 (0.0775,0.0784) | 0.0795 (0.0790,0.0799) |
| Abstract-to-Title Generation | **0.4416** (0.4400,0.4432) | 0.3819 (0.3807,0.3832) | 0.1595 (0.1589,0.1601) | 0.0548 (0.0544,0.0552) | 0.1182 (0.1178,0.1185) | 0.0896 (0.0894,0.0899) | 0.1784 (0.1778,0.1791) | 0.4160 (0.4146,0.4173) | 0.4132 (0.4116,0.4147) |
| Average | **0.2160** | 0.1905 | - | 0.0457 | 0.1160 | 0.1079 | 0.1281 | 0.1828 | 0.1700 |
| **BERTScore** | | | | | | | | | |
| Patient QA | **0.8716** (0.8712,0.8719) | 0.8625 (0.8622,0.8627) | N/A | 0.7890 (0.7886,0.7893) | 0.8373 (0.8370,0.8375) | 0.8394 (0.8392,0.8397) | 0.8398 (0.8395,0.8401) | 0.8463 (0.8460,0.8466) | 0.8393 (0.8390,0.8396) |
| Public Eye Health QA | **0.8480** (0.8479,0.8482) | 0.8381 (0.8380,0.8383) | N/A | 0.7756 (0.7755,0.7757) | 0.8234 (0.8233,0.8235) | 0.8247 (0.8245,0.8248) | 0.8242 (0.8241,0.8244) | 0.8235 (0.8229,0.8241) | 0.8185 (0.8180,0.8191) |
| Patient Consultation | **0.8661** (0.8659,0.8664) | 0.8539 (0.8537,0.8542) | 0.8470 (0.8468,0.8473) | 0.7935 (0.7933,0.7938) | 0.8388 (0.8386,0.8390) | 0.8409 (0.8407,0.8410) | 0.8395 (0.8393,0.8396) | 0.8438 (0.8436,0.8440) | 0.8330 (0.8325,0.8336) |
| Treatment Planning | 0.8251 (0.8248,0.8255) | **0.8229** (0.8227,0.8231) | 0.8090 (0.8083,0.8098) | 0.7725 (0.7717,0.7732) | 0.8061 (0.8053,0.8068) | 0.8085 (0.8083,0.8088) | 0.8194 (0.8192,0.8196) | 0.8139 (0.8137,0.8142) | 0.8017 (0.8015,0.8019) |
| Abstract-to-Title Generation | **0.8898** (0.8895,0.8900) | 0.8786 (0.8784,0.8788) | 0.8477 (0.8475,0.8479) | 0.7982 (0.7977,0.7987) | 0.8340 (0.8339,0.8342) | 0.8185 (0.8183,0.8187) | 0.8539 (0.8537,0.8541) | 0.8834 (0.8832,0.8836) | 0.8796 (0.8792,0.8800) |
| Average | **0.8601** | 0.8512 | - | 0.7858 | 0.8279 | 0.8264 | 0.8354 | 0.8422 | 0.8344 |

*The best scores are highlighted in bold, while the second-best scores are underlined. We performed 10,000 rounds of bootstrap resampling, each sampled to match the size of the original test set. Scores are reported as “Mean (95% CI),” with the best mean scores highlighted in bold.

*Eye-Llama was not evaluated on AAO and iCliniq due to potential training data overlap.

**Extended Data Table 3 | Definitions and counts of eight entity types in visual acuity extraction task.**

| Entity Type | Description | Count | |
|---|---|---|---|
| | | Train | Test |
| VAODDISTCC | Distance visual acuity measurement for the RIGHT eye with correction. | 24 | 42 |
| VAODDISTCCPH | Distance visual acuity measurement for the RIGHT eye with correction and with pinhole. | 2 | 2 |
| VAODDISTSC | Distance visual acuity measurement for the RIGHT eye without correction. | 14 | 31 |
| VAODDISTSCPH | Distance visual acuity measurement for the RIGHT eye without correction and with pinhole. | 3 | 6 |
| VAOSDISTCC | Distance visual acuity measurement for the LEFT eye with correction. | 23 | 47 |
| VAOSDISTCCPH | Distance visual acuity measurement for the LEFT eye with correction and with pinhole (only valid if VAOSDISTCC is present). | 2 | 5 |
| VAOSDISTSC | Distance visual acuity measurement for the LEFT eye without correction. | 15 | 25 |
| VAOSDISTSCPH | Distance visual acuity measurement for the LEFT eye without correction and with pinhole. | 3 | 6 |

*A total of 100 clinical notes were used for training and 200 for testing. The total number of entities was 86 in the training set and 164 in the test set.

**Extended Data Table 4 | Evaluation in Visual Acuity Extraction Task Based on Manually Annotated Real Clinical Notes.**

| Setting | Model | Strict | | | Relaxed | | |
|---|---|---|---|---|---|---|---|
| | | **Precision** | **Recall** | **F1 Score** | **Precision** | **Recall** | **F1 Score** |
| Zero-shot (Open-weight) | LEME-SFT (**Ours**) | **0.5722** | **0.6768** | **0.6201** | **0.6289** | 0.7439 | 0.6816 |
| | LEME-DPO (**Ours**) | 0.4641 | 0.5915 | 0.5201 | 0.6124 | **0.7805** | **0.6863** |
| | LLaMA-3 (70B) | 0.3746 | 0.6646 | 0.4791 | 0.4364 | 0.7744 | 0.5582 |
| | Eye-LLaMA | 0.0187 | 0.0793 | 0.0303 | 0.0259 | 0.1098 | 0.0419 |
| Fine-tuning | LEME-SFT (**Ours**) | 0.8519 | **0.7012** | **0.7692** | **0.8906** | 0.6951 | 0.7808 |
| | LEME-DPO (**Ours**) | **0.8583** | 0.6646 | 0.7491 | 0.8815 | **0.7256** | **0.7960** |
| | LLaMA-3 (70B) | 0.7943 | 0.6829 | 0.7344 | 0.8099 | 0.7012 | 0.7516 |

*The best scores are highlighted in bold. A manually annotated set of 200 private clinical notes was used as the test set. For fine-tuning, 100 clinical notes were used, and the models were trained using LoRA[54].

**Extended Data Table 5 |** Entity type–level analysis of visual acuity extraction performance.

| | LEME-SFT | | | LEME-DPO | | | LLaMA-3 (70B) | | |
|---|---|---|---|---|---|---|---|---|---|
| **Type** | **Precision** | **Recall** | **F1** | **Precision** | **Recall** | **F1** | **Precision** | **Recall** | **F1** |
| VAODDISTCC | 0.872 | 0.810 | 0.840 | 0.861 | 0.738 | 0.795 | 0.800 | 0.762 | 0.780 |
| VAODDISTCCPH | 0.400 | 1.000 | 0.571 | 0.400 | 1.000 | 0.571 | 0.400 | 1.000 | 0.571 |
| VAODDISTSC | 0.909 | 0.645 | 0.755 | 0.913 | 0.677 | 0.778 | 0.870 | 0.645 | 0.741 |
| VAODDISTSCPH | 0.667 | 0.667 | 0.667 | 0.800 | 0.667 | 0.727 | 0.667 | 0.667 | 0.667 |
| VAOSDISTCC | 0.889 | 0.681 | 0.771 | 0.848 | 0.596 | 0.700 | 0.825 | 0.702 | 0.759 |
| VAOSDISTCCPH | 1.000 | 0.800 | 0.889 | 1.000 | 0.800 | 0.889 | 1.000 | 0.600 | 0.750 |
| VAOSDISTSC | 0.833 | 0.600 | 0.698 | 0.889 | 0.640 | 0.744 | 0.789 | 0.600 | 0.682 |
| VAOSDISTSCPH | 0.800 | 0.667 | 0.727 | 1.000 | 0.500 | 0.667 | 0.600 | 0.500 | 0.545 |
| Total | 0.852 | 0.701 | 0.769 | 0.858 | 0.665 | 0.749 | 0.794 | 0.683 | 0.734 |

**Extended Data Table 6 l** Evaluation of benchmark performance on multiple-choice question answering.

| Task | Ours | | Publicly Available Models | | | | | Proprietary Models | |
|---|---|---|---|---|---|---|---|---|---|
| | LEME-SFT (70B) | LEME-DPO (70B) | EYE-LLaMA (7B) | MediTron (70B) | LLaMA-2 (7B) | LLaMA-2 (70B) | LLaMA-3 (70B) | GPT-4o | o1-mini |
| BCSC | 0.7728<br>(0.7676,0.7781) | <u>0.8060</u><br>(0.8013,0.8108) | N/A | 0.5685<br>(0.5626,0.5743) | 0.3436<br>(0.3375,0.3496) | 0.4034<br>(0.3978,0.4090) | 0.7683<br>(0.7632,0.7735) | 0.8833<br>(0.8791,0.8875) | 0.8042<br>(0.7993,0.8091) |

*The best scores are highlighted in bold, while the second-best scores are underlined. We performed 10,000 rounds of bootstrap resampling, each sampled to match the size of the original test set. Scores are reported as “Mean ± SD (95% CI),” with the best mean scores highlighted in bold.

*Eye-Llama was not evaluated on BCSC due to potential training data overlap.

**Extended Data Table 7 |** Details of ophthalmology corpora: sources, licenses, and segment counts.

| Corpus | Original Source | Download (Pre-processed Version) | License | Ophthalmology Segments (#) |
|---|---|---|---|---|
| Clinical Guidelines | the National Institute for Health and Care Excellence (NICE), Centers for Disease Control and Prevention (CDC), World Health Organization (WHO), Canadian Medical Association (CMA), Cancer Care Ontario (CCO), Strategy for Patient-Oriented Research (SPOR), International Committee of the Red Cross (ICRC), PubMed, and WikiDoc | https://huggingface.co/datasets/epfl-llm/guidelines | Common Crawl | 3,435 |
| Textbooks | StatPearls (https://www.statpearls.com/) | https://github.com/Teddy-XiongGZ/MIRAGE | CC BY-NC 4.0 | 12,919 |
| Case reports | Open-access articles from 105 ophthalmology journals indexed in PubMed Central:<br>Cont Lens Anterior Eye, Retin Cases Brief Rep, Ophthalmic Surg Lasers Imaging Retina, J Optom, Opt Express, Ophthalmology, J Binocul Vis Ocul Motil, Ophthalmic Surg, Acta Ophthalmol, Turk J Ophthalmol, Vis Neurosci, Arq Bras Oftalmol, Ophthalmol Glaucoma, Korean J Ophthalmol, BMJ Open Ophthalmol, Ophthalmic Surg Lasers Imaging, Ann Otol Rhinol Laryngol, Open Ophthalmol J, Transl Vis Sci Technol, Ophthalmic Genet, CLAO J, Zhonghua Yan Ke Za Zhi, Adv Ophthalmic Plast Reconstr Surg, Optom Vis Sci, Doc Ophthalmol, Strabismus, Ophthalmologica, J Ocul Pharmacol Ther, Case Rep Ophthalmol Med, Surv Ophthalmol, J Glaucoma, Oman J Ophthalmol, Dev Ophthalmol, Asia Pac J Ophthalmol (Phila), Digit J Ophthalmol, Int Ophthalmol Clin, Eye (Lond), Arch Soc Esp Oftalmol, Trans Am Ophthalmol Soc, Indian J Ophthalmol, Trans Pa Acad Ophthalmol Otolaryngol, Am J Ophthalmol, Compr Ophthalmol Update, Br J Ophthalmol, Clin Exp Optom, Clin Exp Ophthalmol, Eur J Ophthalmol, J Opt Soc Am A Opt Image Sci Vis, J Fr Ophtalmol, Vision Res, Case Rep Ophthalmol , Ophthalmic Res, Annu Rev Vis Sci, Semin Ophthalmol, Ophthalmologie, Ophthalmol Retina, Jpn J Ophthalmol, Multisens Res, Cornea, JAMA Ophthalmol, Nepal J Ophthalmol, Mol Vis, Cutan Ocul Toxicol, Curr Eye Res, Expert Rev Ophthalmol, Int Ophthalmol, J Pediatr Ophthalmol Strabismus, Ophthalmol Sci, J Biomed Opt, Cesk Slov Oftalmol, Ocul Immunol Inflamm, Can J Ophthalmol, Graefes Arch Clin Exp Ophthalmol, J Curr Glaucoma Pract, Klin Monbl Augenheilkd, Middle East Afr J Ophthalmol, Orbit, Ophthalmic Epidemiol, J Vis, Vet Ophthalmol, J Neuroophthalmol, BMC Ophthalmol, Rom J Ophthalmol, Prog Retin Eye Res, Clin Ophthalmol, Arch Ophthalmol, Invest Ophthalmol Vis Sci, J Ophthalmic Vis Res, Eye Sci, J AAPOS, Eye Contact Lens, J Cataract Refract Surg, Retina, Ophthalmic Plast Reconstr Surg, Exp Eye Res, Vestn Oftalmol, Opt Lett, Ophthalmic Physiol Opt, Optometry, Ocul Surf, Curr Opin Ophthalmol, J Clin Exp Ophthalmol, J Refract Surg, Appl Opt, Am J Ophthalmol Case Rep | https://huggingface.co/datasets/zhengyun21/PMC-Patients | CC BY-NC-SA 4.0 | 5,745 |

# Supplementary Information

## Section 1 – Qualitative analysis of model reasoning on BCSC

To qualitatively assess reasoning quality, an ophthalmologist reviewed model outputs from LEME-SFT and LLaMA-3 70B, selecting the better response or declaring a draw. Human-written answers were provided as references to guide the assessment. The evaluation considered factual accuracy, explanation of incorrect options, and overall clarity, and was conducted on a 25-question subset of the BCSC dataset. As the two models achieved similar quantitative performance on the full dataset (LEME-SFT: 0.7728; LLaMA-3 70B: 0.7683), this comparison reflects reasoning quality with minimal influence from overall accuracy differences.

The reviewed cases are provided in **Supplementary Table 1**. Among these, LEME-SFT was judged superior in 13 cases, equivalent in 10, and inferior in only 2. Despite comparable accuracy, the evaluation revealed that LEME consistently produced more accurate, comprehensive, and interpretable reasoning. In contrast, LLaMA-3 often failed to justify incorrect options or included flawed logic, even when the final answer was correct. These findings highlight the impact of supervised fine-tuning in generating not only accurate but also explainable outputs, a critical requirement for medical applications.

**Supplementary Table 1** | Head-to-head comparison of reasoning quality between LEME-SFT and LLaMA-3 70B on BCSC questions. Inaccurate or suboptimal responses are highlighted in red. Cases where LEME-SFT outperformed LLaMA-3 70B (Questions 3, 6, 10–13, 15, 18–20, 23–25) are shown in green, while cases favoring LLaMA-3 70B (Questions 4 and 16) are shown in purple. Some response segments were truncated for readability.

| | |
|---|---|
| **Question #1**: A 34-year-old woman in her 34th week of pregnancy develops pre-eclampsia. She is monitored | **LLaMA-3**: **Summary of available information:**<br>* The patient is a 34-year-old woman in her 34th week of pregnancy with pre-eclampsia. |
| successfully until she presents to the emergency department with acute-onset headache, altered mental status, and complete vision loss in both eyes. Magnetic resonance imaging (MRI) shows increased fluid-attenuated inversion recovery (FLAIR) signal in the bilateral occipital lobes. After she is stabilized, how should the ophthalmologist counsel her family regarding her visual prognosis?<br><br>A: She will regain some vision and have a persistent homonymous hemianopsia<br>B: She will likely recover most or all of her vision<br>C: She will be permanently blind<br>D: She will regain some vision but will be unable to detect color | * She presents to the emergency department with acute-onset headache, altered mental status, and complete vision loss in both eyes.<br>* MRI shows increased fluid-attenuated inversion recovery (FLAIR) signal in the bilateral occipital lobes.<br><br>**Analysis:**<br>The patient's symptoms and MRI findings suggest posterior reversible encephalopathy syndrome (PRES), a condition that can occur in the setting of pre-eclampsia. PRES is characterized by reversible vasogenic edema in the brain, often affecting the occipital lobes, which can cause vision loss.<br><br>**Visual prognosis:**<br>Given the diagnosis of PRES, the patient's visual prognosis is generally good. The condition is reversible, and vision often returns to normal or near-normal after treatment and resolution of the underlying condition.<br><br>**LEME**: The patient is a 34-year-old woman in her 34th week of pregnancy who has developed pre-eclampsia and is now presenting with symptoms of acute-onset headache, altered mental status, and complete vision loss in both eyes. MRI findings show increased FLAIR signal in the bilateral occipital lobes, which suggests posterior reversible encephalopathy syndrome (PRES), a condition often associated with pre-eclampsia. PRES typically causes reversible brain edema, and with appropriate management, many patients experience significant recovery of neurological symptoms, including vision. Therefore, the most likely prognosis for her vision is that she will recover most or all of her vision. |
| **Question #2**: A child with uncorrected infantile esotropia and manifest latent nystagmus has a face turn to the left. This head posture is most likely related to what?<br><br>A: fusing gaze position<br>B: null point | **LLaMA-3**: **Summary of available information:**<br>* The patient is a child with uncorrected infantile esotropia, which means they have a type of strabismus (eye misalignment) where the eyes turn inward.<br>* The child also has manifest latent nystagmus, which is a type of involuntary eye movement.<br>* The child has a face turn to the left, which means they are turning their head to the left to compensate for their eye misalignment.<br><br>**Analysis:** |
| C: 'pinhole' effect with uncorrected refractive error<br>D: temporal suppression scotoma | * Option A, fusing gaze position, is unlikely to be the correct answer because the child has uncorrected esotropia, which means their eyes are not aligned properly, making it difficult to fuse images.<br>* Option C, 'pinhole' effect with uncorrected refractive error, is also unlikely because the question doesn't mention any refractive error, and the 'pinhole' effect is related to the way light enters the eye, not head posture.<br>* Option D, temporal suppression scotoma, is a possibility, but it's more related to visual field defects, which isn't directly related to head posture.<br><br>**Conclusion:**<br>The most likely explanation for the child's head turn to the left is to compensate for their eye misalignment, specifically to find a position where their eyes are more aligned, which is known as the null point. This is a common phenomenon in patients with strabismus.<br><br>**LEME**: Infantile esotropia is a form of strabismus where one or both eyes turn inward. Manifest latent nystagmus is a type of involuntary eye movement that occurs when the eyes are not aligned. A face turn to the left in a child with these conditions is likely an attempt to adopt a head posture that minimizes the nystagmus and stabilizes vision. This is often achieved by finding a "null point," where the nystagmus is minimized and the vision is more stable. The other options do not directly relate to the head posture adopted to manage nystagmus. |
| **Question #3**: What is the likely cause of increased intraocular pressure in the setting of chronic vitreous hemorrhage?<br><br>A: sickling of red blood cells<br>B: Schwartz-Matsuo syndrome<br>C: 'ghost cell' degeneration of red blood cells<br>D: synchysis scintillans | **LLaMA-3**: **Summary of available information:**<br>The question is asking about the likely cause of increased intraocular pressure in the setting of chronic vitreous hemorrhage. Vitreous hemorrhage refers to bleeding into the vitreous gel that fills the center of the eye. Chronic vitreous hemorrhage means that the bleeding has been ongoing for a prolonged period.<br><br>**Analysis:**<br>Increased intraocular pressure in the setting of chronic vitreous hemorrhage can be caused by several factors. Among the options, we need to identify the most likely cause.<br><br>**Option A: Sickling of red blood cells** |

Sickling of red blood cells is a characteristic of sickle cell disease, which can cause vaso-occlusion and tissue damage. While sickling can occur in the eye, it is not a direct cause of increased intraocular pressure in chronic vitreous hemorrhage.

**Option B: Schwartz-Matsuo syndrome**
Schwartz-Matsuo syndrome is a rare condition characterized by a sudden, severe vitreous hemorrhage. While it is related to vitreous hemorrhage, it is not a cause of increased intraocular pressure.

**Option C: 'Ghost cell' degeneration of red blood cells**
Ghost cells are degenerated red blood cells that can accumulate in the vitreous humor. As they break down, they can release hemoglobin, which can be converted to hemosiderin. Hemosiderin can stimulate the production of fibrovascular tissue, leading to traction on the retina and increased intraocular pressure.

**Option D: Synchysis scintillans**
Synchysis scintillans is a condition characterized by the accumulation of cholesterol crystals in the vitreous humor, often seen in chronic vitreous hemorrhage. While it is associated with vitreous hemorrhage, it is not a direct cause of increased intraocular pressure.

**Conclusion:**
Based on the analysis, the most likely cause of increased intraocular pressure in the setting of chronic vitreous hemorrhage is the 'ghost cell' degeneration of red blood cells, which can lead to the production of fibrovascular tissue and traction on the retina.

**LEME**: To determine the likely cause of increased intraocular pressure in the setting of chronic vitreous hemorrhage, we need to consider the pathophysiology of each option:

A: Sickling of red blood cells - This is related to sickle cell disease and not typically associated with chronic vitreous hemorrhage.
B: Schwartz-Matsuo syndrome - This syndrome involves the blockage of the trabecular meshwork by photoreceptor outer segments, typically associated with retinal detachment, not chronic vitreous hemorrhage.

| | |
|---|---|
| | C: 'Ghost cell' degeneration of red blood cells - Ghost cells are degenerated red blood cells that can clog the trabecular meshwork, leading to increased intraocular pressure in the setting of chronic vitreous hemorrhage.<br>D: Synchysis scintillans - This is a condition characterized by cholesterol crystals in the vitreous and is not typically associated with increased intraocular pressure due to chronic vitreous hemorrhage.<br><br>Given this analysis, the most likely cause of increased intraocular pressure in the setting of chronic vitreous hemorrhage is the 'ghost cell' degeneration of red blood cells. |
| **Question #4**: A 12-year-old boy is diagnosed with well-controlled, intermittent exotropia. What is stereopsis testing most likely to reveal?<br><br>A: no stereopsis because of temporal hemiretinal suppression<br>B: no stereopsis because of diplopia<br>C: monofixation syndrome<br>D: excellent stereopsis | **LLaMA-3**: **Summary of available information:**<br>* The patient is a 12-year-old boy.<br>* He has been diagnosed with well-controlled, intermittent exotropia.<br><br>**Understanding the diagnosis:**<br>* Exotropia is a type of strabismus, where the eyes are misaligned and point outward.<br>* Intermittent exotropia means that the misalignment occurs occasionally, but not always.<br>* Well-controlled implies that the exotropia is not severe and can be managed with treatment or exercises.<br><br>**Stereopsis testing:**<br>* Stereopsis is the ability to perceive depth and distance using both eyes together.<br>* In a patient with exotropia, stereopsis testing can help determine the extent of binocular vision and depth perception.<br><br>**Analyzing the options:**<br>* A: No stereopsis because of temporal hemiretinal suppression is unlikely, as there is no indication of hemiretinal suppression in the question.<br>* B: No stereopsis because of diplopia is also unlikely, as diplopia (double vision) is not mentioned in the question.<br>* C: Monofixation syndrome is a condition where one eye is used for fixation, and the other eye is suppressed. While possible, it's not the most likely answer given the diagnosis of intermittent exotropia. |

| | * D: Excellent stereopsis is a plausible answer, as the patient has well-controlled intermittent exotropia, which suggests that their binocular vision and depth perception may be relatively preserved.<br><br>## Answer: D: excellent stereopsis<br><br>**Comment**: (1) Poor reasoning. Just the fact that diplopia isn't mentioned in the question doesn't mean that it couldn't be the answer. (2) Nice to mention “well-controlled intermittent exotropia.”<br><br>**LEME**: Exotropia is a form of strabismus where one or both eyes turn outward. In intermittent exotropia, the eyes are not always misaligned and can sometimes be controlled. Stereopsis refers to the ability to perceive depth based on the different views from both eyes. In cases of intermittent exotropia, the brain may suppress the image from one eye to avoid double vision (diplopia), which can lead to a lack of stereopsis. Temporal hemiretinal suppression occurs when the brain ignores the visual input from the temporal (outer) half of the retina of one eye to avoid confusion and double vision. Monofixation syndrome is a condition where one eye is used for fixation while the other eye is suppressed, often seen in strabismus. Given that the patient has well-controlled intermittent exotropia, they are likely to have some degree of stereopsis because their eyes can align properly at times. However, the most likely scenario is that they have reduced or no stereopsis due to the brain's suppression mechanism to avoid diplopia.<br><br>## Answer: B<br><br>**Comment**: This is not the most likely scenario. In the early stages of the disorder, the deviation is usually larger for distance viewing than for near. Thus, sensory testing at near is typically excellent, including stereopsis and fusion on Worth 4 Dot testing. |
| --- | --- |
| **Question #5**: A 12-year-old patient with juvenile idiopathic arthritis (JIA)–associated chronic anterior uveitis (CAU) presents for her regular follow-up appointment. She has been a patient at this clinic for 7 years. Her CAU has been controlled, but she has | **LLaMA-3**: **Summary of available information:**<br>* The patient is 12 years old and has juvenile idiopathic arthritis (JIA) associated with chronic anterior uveitis (CAU).<br>* The patient has been followed up at the clinic for 7 years.<br>* The CAU has been controlled, but the patient requires 3 times daily topical steroids to maintain control. |
| required 3 times daily topical steroids in order for the inflammation to remain quiet. What finding can be expected on examination?<br><br>A: fibrinous anterior chamber reaction<br>B: vitritis<br>C: cataract<br>D: hypopyon | **Analysis:**<br>Given the patient's history of chronic anterior uveitis and the need for long-term topical steroid use, we can expect a complication related to steroid use.<br><br>**Expected finding:**<br>The most likely finding on examination is a cataract, which is a common complication of long-term steroid use.<br><br>**LEME**: Juvenile idiopathic arthritis (JIA)-associated chronic anterior uveitis (CAU) is a common condition in pediatric patients with JIA. The chronic use of topical steroids, while effective in controlling inflammation, can lead to complications. One common complication of chronic steroid use in the eye is the development of cataracts. Fibrinous anterior chamber reaction, vitritis, and hypopyon are not typically associated with chronic steroid use but are more indicative of active inflammation or infection. Given the patient's history of well-controlled CAU with chronic steroid use, the most likely finding on examination would be a cataract. |
| **Question #6**: A 14-year-old girl has 10 prism diopters of exodeviation in upgaze and 10 prism diopters of esodeviation in downgaze. Examination of her extraocular movements is most likely to reveal overaction of what extraocular muscles?<br><br>A: inferior rectus muscles<br>B: superior rectus muscles<br>C: superior oblique muscles<br>D: inferior oblique muscles | **LLaMA-3**: **Summary of available information:**<br>* The patient is a 14-year-old girl.<br>* She has 10 prism diopters of exodeviation in upgaze (meaning her eyes deviate outward when looking up).<br>* She has 10 prism diopters of esodeviation in downgaze (meaning her eyes deviate inward when looking down).<br><br>**Analysis:**<br>* Exodeviation in upgaze suggests that the muscles that normally move the eyes inward (adductors) are weak or inhibited, allowing the eyes to deviate outward.<br>* Esodeviation in downgaze suggests that the muscles that normally move the eyes outward (abductors) are weak or inhibited, allowing the eyes to deviate inward.<br>* Considering both findings together, it is likely that the inferior oblique muscles are overacting, as they are responsible for elevating and abducting the eye (moving it upward and outward).<br><br>**Comment**: The question is asking about overreaction, the reasoning not really to the point. |
| | **LEME**: The question describes a 14-year-old girl with a specific pattern of eye deviation: exodeviation (outward deviation) in upgaze and esodeviation (inward deviation) in downgaze. This pattern suggests an issue with the muscles that control vertical and torsional movements of the eyes. The inferior oblique muscles are responsible for elevating the eye and turning it outward, while the superior oblique muscles are responsible for depressing the eye and turning it inward. Given the pattern of deviations described, overaction of the inferior oblique muscles would cause the eye to deviate outward in upgaze and inward in downgaze. Therefore, the most likely cause is overaction of the inferior oblique muscles. |
| **Question #7**: A 1-month-old baby is admitted to the intensive care unit (ICU) with bulging fontanelles, a spiral fracture of the femur and multiple skin bruises. What would be the most appropriate time for a fundus examination?<br><br>A: as soon as possible<br>B: on discharge from the ICU<br>C: as an outpatient<br>D: within 2 to 3 weeks | **LLaMA-3**: **Summary of available information:**<br>* The patient is a 1-month-old baby.<br>* The baby is admitted to the ICU with:<br>+ Bulging fontanelles (a sign of increased intracranial pressure).<br>+ A spiral fracture of the femur (a type of fracture that is often associated with non-accidental injury).<br>+ Multiple skin bruises (another sign of possible non-accidental injury).<br><br>**Analysis:**<br>Given the symptoms and signs of possible non-accidental injury, it is essential to rule out any underlying conditions that may be contributing to these findings. A fundus examination is a crucial step in evaluating for retinal hemorrhages, which are often seen in cases of shaken baby syndrome.<br><br>**Conclusion:**<br>The most appropriate time for a fundus examination would be as soon as possible, as it may provide valuable information about the underlying cause of the baby's symptoms and inform further management.<br><br>**LEME**: The question describes a 1-month-old baby with bulging fontanelles, a spiral fracture of the femur, and multiple skin bruises, which are concerning signs of possible child abuse or shaken baby syndrome. A fundus examination is important to check for retinal hemorrhages, which can be a sign of shaken baby syndrome. Given the severity of the symptoms and the need for a thorough assessment to rule out or confirm abuse, the most appropriate time for a fundus examination would be as soon as possible to ensure prompt diagnosis and intervention. |
| **Question #8**: A 20-year-old patient with a history of seizures presents for a general eye examination. During the examination, the patient loses consciousness, and on subsequent electroencephalogram (EEG) testing with her neurologist is found to have abnormal activity in both hemispheres. What type of seizure did the patient experience in the ophthalmologist's office?<br><br>A: partial<br>B: simple<br>C: generalized<br>D: complex | **LLaMA-3**: * The patient has a history of seizures.<br>* The patient loses consciousness during the eye examination.<br>* The EEG testing shows abnormal activity in both hemispheres.<br><br>From the information, we can infer that the patient experienced a seizure that affected both sides of the brain, as indicated by the abnormal activity in both hemispheres on the EEG. This suggests that the seizure was not limited to a specific area of the brain, but rather was more widespread. Given this, the correct answer is: C: generalized<br><br>**LEME**: The question describes a patient with a history of seizures who loses consciousness during an eye examination. The subsequent EEG testing reveals abnormal activity in both hemispheres of the brain. This suggests that the seizure affected both sides of the brain simultaneously, which is characteristic of a generalized seizure. Partial seizures, also known as focal seizures, typically involve only one hemisphere. Simple and complex are subtypes of partial seizures, with simple partial seizures involving no loss of consciousness and complex partial seizures involving impaired consciousness but still typically affecting only one hemisphere. |
| **Question #9**: A 21-year-old woman develops asymptomatic, hyperemic disc edema in the right eye with telangiectatic changes. Her visual acuity is 20/20, and she has a trace relative afferent pupillary defect (RAPD). What is most likely responsible for this examination finding?<br><br>A: nocturnal hypotension with crowded optic disc<br>B: type 1 diabetes<br>C: Leber hereditary optic neuropathy<br>D: optic nerve glioma | **LLaMA-3**: **Summary of available information:**<br>* The patient is a 21-year-old woman.<br>* She has asymptomatic, hyperemic disc edema in the right eye.<br>* There are telangiectatic changes present.<br>* Her visual acuity is 20/20, which is normal.<br>* She has a trace relative afferent pupillary defect (RAPD), which suggests some degree of optic nerve dysfunction.<br><br>**Analysis:**<br>Based on the symptoms and signs, we can rule out option D (optic nerve glioma) as it would likely cause more significant visual impairment and a more pronounced RAPD. Option B (type 1 diabetes) is also unlikely as it would typically cause diabetic retinopathy, not isolated optic disc edema. The presence of telangiectatic changes and hyperemic disc edema suggests a possible mitochondrial disorder. Leber hereditary optic neuropathy (LHON) is a mitochondrial disorder that can cause optic disc edema, telangiectatic changes, and a RAPD. It typically affects young adults, which fits the patient's age. |
| | Option A (nocturnal hypotension with crowded optic disc) is possible, but it would not typically cause telangiectatic changes.<br><br>**Answer:** C: Leber hereditary optic neuropathy ##<br><br>**LEME**: The question describes a 21-year-old woman with asymptomatic, hyperemic disc edema in the right eye, telangiectatic changes, 20/20 visual acuity, and a trace relative afferent pupillary defect (RAPD). The key points to consider are the presence of disc edema, hyperemia, and telangiectasia, along with the absence of significant vision loss.<br><br>- Nocturnal hypotension with a crowded optic disc (A) can cause optic disc edema due to reduced blood flow to the optic nerve during sleep.<br>- Type 1 diabetes (B) can lead to various ocular complications, but it is less likely to present with isolated optic disc edema and telangiectasia without other systemic signs.<br>- Leber hereditary optic neuropathy (C) typically presents with painless, subacute central vision loss and does not usually cause telangiectasia or hyperemic disc edema.<br>- Optic nerve glioma (D) can cause optic disc edema and visual field defects, but it is less likely to present with telangiectasia and typically causes more significant vision loss.<br><br>Given the combination of findings, nocturnal hypotension with a crowded optic disc is the most likely cause.<br><br>## Answer: A |
| **Question #10**: A 22-year-old man presents with night blindness, constricted visual fields, posterior subcapsular cataracts, and cystoid macular edema. What is the most likely diagnosis?<br><br>A: cone dystrophy, B: cancer-associated retinopathy (CAR), C: | **LLaMA-3**: **Summary of available information:**<br>* The patient is a 22-year-old man.<br>* He presents with:<br>+ Night blindness (difficulty seeing in low light)<br>+ Constricted visual fields (narrowing of the field of vision)<br>+ Posterior subcapsular cataracts (a type of cataract that forms at the back of the lens)<br>+ Cystoid macular edema (fluid accumulation in the macula, the part of the retina responsible for central vision)<br><br>**Analysis:** |

| retinitis pigmentosa, D: vitamin A deficiency | Based on the symptoms, we can narrow down the possible causes. Night blindness and constricted visual fields are common in retinal disorders. Posterior subcapsular cataracts are often associated with retinal diseases. Cystoid macular edema is a complication of various retinal conditions.<br><br>**Differential diagnosis:**<br>* Cone dystrophy: a rare inherited disorder affecting the cone cells in the retina, leading to vision loss. While it can cause night blindness and constricted visual fields, it is less likely to cause posterior subcapsular cataracts and cystoid macular edema.<br>* Cancer-associated retinopathy (CAR): a paraneoplastic syndrome that can cause retinal degeneration, but it is typically associated with cancer, which is not mentioned in the question.<br>* Retinitis pigmentosa: a group of inherited disorders that affect the retina, causing progressive vision loss. It is a good fit for the symptoms, as it can cause night blindness, constricted visual fields, posterior subcapsular cataracts, and cystoid macular edema.<br>* Vitamin A deficiency: can cause night blindness, but it is less likely to cause the other symptoms mentioned.<br><br>**LEME**: The patient's symptoms include night blindness, constricted visual fields, posterior subcapsular cataracts, and cystoid macular edema. Let's analyze each option:<br><br>A: Cone dystrophy typically presents with decreased central vision and color vision defects, but not necessarily night blindness or constricted visual fields.<br>B: Cancer-associated retinopathy (CAR) can cause visual field defects and vision loss, but it is less commonly associated with night blindness and posterior subcapsular cataracts.<br>C: Retinitis pigmentosa is characterized by night blindness, constricted visual fields, and can also be associated with cataracts and cystoid macular edema.<br>D: Vitamin A deficiency can cause night blindness, but it is less likely to cause constricted visual fields, posterior subcapsular cataracts, and cystoid macular edema. |
|---|---|
| **Question #11**: A 25-year-old man presents with episodes of sharp, ice-pick-like right eye pain, right eye redness and tearing. In the differential for epiphora what is the most likely diagnosis? | **LLaMA-3**: **Summary of available information:**<br>* The patient is a 25-year-old man.<br>* He presents with episodes of sharp, ice-pick-like right eye pain.<br>* He also experiences right eye redness and tearing.<br>* The question asks for the most likely diagnosis in the differential for epiphora (excessive tearing). |
| A: ectropion, B: canaliculitis, C: cluster headache, D: acquired nasolacrimal duct obstruction | **Analysis:**<br>* The symptoms described, particularly the sharp, ice-pick-like pain, are consistent with a headache disorder.<br>* The presence of right eye redness and tearing suggests that the headache disorder may be related to the eye.<br>* Cluster headache is a type of headache disorder that is known to cause eye symptoms, including tearing and redness.<br><br>**Conclusion:**<br><br>Based on the symptoms described, the most likely diagnosis is a cluster headache.<br><br>## Answer: C: cluster headache<br><br>***Comment**: Correct but not an entirely complete/satisfying answer. Model does not acknowledge that this specific presentation is consistent with cluster headache because cluster headaches are characterizzed by attacks of intense pain that leaves the patient unable to rest, ipsilateral tearing, conjunctival injection, rhinorrhea, and postganglionic Horner syndrome.*<br><br>**LEME**: The patient is a 25-year-old man experiencing sharp, ice-pick-like pain in the right eye, along with redness and tearing. The differential diagnosis for epiphora (excessive tearing) includes conditions that can cause similar symptoms.<br><br>- Ectropion (A) is a condition where the eyelid turns outward, which is less likely to cause sharp pain.<br>- Canaliculitis (B) is an infection of the canaliculi, which can cause tearing and discharge but typically not sharp, ice-pick-like pain.<br>- Cluster headache (C) can cause severe unilateral pain and autonomic symptoms like tearing and redness, fitting the description of sharp, ice-pick-like pain.<br>- Acquired nasolacrimal duct obstruction (D) usually causes chronic tearing and discharge but not typically sharp, acute pain. |
| **Question #12**: A 25-year-old woman was recently diagnosed with pars planitis. Her visual acuity is 20/20 OD, | **LLaMA-3**: **Summary of available information:**<br>* The patient is a 25-year-old woman with a diagnosis of pars planitis.<br>* Her visual acuity is 20/20 in the right eye (OD) and 20/50 in the left eye (OS). |
| 20/50 OS. The examination reveals vitreous cells and cystoid macular edema (CME) in the left eye. What is the most appropriate patient management plan?<br><br>A: oral methotrexate<br>B: pars plana vitrectomy OS<br>C: argon laser treatment of the peripheral retina<br>D: sub-Tenon corticosteroid injection OS | * The examination reveals vitreous cells and cystoid macular edema (CME) in the left eye.<br><br>**Analysis:**<br>Pars planitis is a type of intermediate uveitis, which is an inflammatory condition affecting the middle layer of the eye. The presence of vitreous cells and CME in the left eye suggests active inflammation. The patient's decreased visual acuity in the left eye (20/50) indicates that the inflammation is affecting her vision.<br><br>**Management plan:**<br>The goal of management is to reduce inflammation and prevent further vision loss. Given the presence of CME, which is a common complication of uveitis, the most appropriate management plan would be to treat the inflammation locally.<br><br>**LEME**: Pars planitis is an intermediate uveitis that can lead to complications such as cystoid macular edema (CME), which can affect vision. The patient's visual acuity is reduced in the left eye (OS), and the presence of vitreous cells and CME indicates active inflammation. The goal of treatment is to reduce inflammation and manage the CME to improve or preserve vision.<br><br>Oral methotrexate (A) is an immunosuppressive agent used for chronic or severe cases of uveitis but is not typically the first line of treatment for pars planitis with CME. Pars plana vitrectomy (B) is a surgical procedure that is usually reserved for severe cases or when there is a non-resolving vitreous hemorrhage or retinal detachment. Argon laser treatment (C) is not a standard treatment for pars planitis. Sub-Tenon corticosteroid injection (D) is a localized treatment that can effectively reduce inflammation and manage CME in the eye.<br><br>Given the information, the most appropriate management plan for this patient with active inflammation and CME in the left eye is a sub-Tenon corticosteroid injection. |
| **Question #13**: A 29-year-old woman with longstanding insulin-dependent diabetes has a 9-pound baby boy. This history increases the risk of what optic nerve condition? | **LLaMA-3**: **Summary of available information:**<br>* The patient is a 29-year-old woman with a history of insulin-dependent diabetes.<br>* She has given birth to a 9-pound baby boy.<br><br>**Relevant medical knowledge:**<br>* Insulin-dependent diabetes is a risk factor for certain optic nerve conditions. |

| A: tilted disk<br>B: morning glory disk anomaly<br>C: optic nerve coloboma<br>D: optic nerve hypoplasia | * A large birth weight (macrosomia) is also associated with an increased risk of optic nerve abnormalities.<br><br>**Analysis:**<br>Based on the patient's history of insulin-dependent diabetes and the birth of a large baby, we need to identify the optic nerve condition that is most likely to be associated with these risk factors.<br><br>**Option analysis:**<br>* A: Tilted disk is a congenital anomaly, but it is not directly related to diabetes or macrosomia.<br>* B: Morning glory disk anomaly is a rare congenital defect, but it is not associated with diabetes or macrosomia.<br>* C: Optic nerve coloboma is a congenital defect, but it is not directly related to diabetes or macrosomia.<br>* D: Optic nerve hypoplasia is a condition where the optic nerve is underdeveloped, and it is associated with maternal diabetes and macrosomia.<br><br>**LEME**: The question describes a 29-year-old woman with insulin-dependent diabetes who has given birth to a 9-pound baby boy. This suggests that the baby is macrosomic, which is a common condition in infants of diabetic mothers due to increased glucose levels during pregnancy. Macrosomia is associated with various congenital anomalies. Among the options provided, optic nerve hypoplasia is known to be associated with maternal diabetes. Optic nerve hypoplasia is a condition where the optic nerve is underdeveloped, and it can be linked to various systemic conditions, including maternal diabetes. |
|---|---|
| **Question #14**: A 30-year-old man presents for evaluation of “one eye turning out” when he is tired. He has experienced this for several years. Initial evaluation demonstrates orthophoria. Upon cover testing, an exodeviation becomes present, although he is able to resume fusion with blinking. How is his exodeviation most accurately described? | **LLaMA-3**: Let's break down the available information:<br>* The patient presents with a complaint of "one eye turning out" when he is tired, suggesting a possible strabismus (eye misalignment) issue.<br>* The initial evaluation demonstrates orthophoria, meaning the eyes are aligned properly when looking at a distant object.<br>* Upon cover testing, an exodeviation (outward deviation of one eye) becomes present, indicating that the eyes are not aligned properly when one eye is covered.<br>* However, the patient is able to resume fusion (align the eyes properly) with blinking, suggesting that the exodeviation is not constant. |
| A: intermittent exotropia with poor control<br>B: intermittent exotropia with good control<br>C: intermittent exotropia with fair control<br>D: exophoria | Based on this information, we can conclude that the patient has an intermittent exotropia, which means the exodeviation is not present all the time but appears under certain conditions (in this case, when tired).<br><br>The key to determining the level of control is the patient's ability to resume fusion with blinking. Since the patient can resume fusion with blinking, it indicates that they have good control over their exodeviation.<br><br>**LEME**: The patient is a 30-year-old man who experiences one eye turning out when he is tired, suggesting a condition related to eye alignment that varies with fatigue. The initial evaluation shows orthophoria, meaning no deviation is present when he is not tired. However, upon cover testing, an exodeviation (outward turning of the eye) becomes apparent, but he can resume normal eye alignment (fusion) with blinking. This indicates that the exodeviation is not constant but rather intermittent. The fact that he can regain alignment with blinking suggests that he has good control over the deviation. Therefore, the most accurate description of his condition is intermittent exotropia with good control. |
| **Question #15**: A 30-year-old man was using an electric power saw on a piece of wood and hit a nail. He noticed a sharp pain in the left eye, with worsening vision over the next 2 days. He was not wearing goggles. He presents with visual acuity of 20/20 and redness and chemosis inferotemporally in his left eye 4 mm from the limbus. The pupil is round and reactive. Dilated examination is remarkable for a localized hemorrhage in the inferior vitreous, but no foreign body is visualized. What is the next appropriate step in management? | **LLaMA-3**: **Summary of available information:**<br>* A 30-year-old man was using an electric power saw and hit a nail, causing a sharp pain in his left eye.<br>* He didn't wear goggles.<br>* He presents with:<br>+ Visual acuity of 20/20 (normal vision).<br>+ Redness and chemosis (swelling) inferotemporally (lower outer area) in his left eye, 4 mm from the limbus (border of the cornea).<br>+ Round and reactive pupil.<br>+ Localized hemorrhage in the inferior vitreous (lower part of the gel-like substance in the eye) on dilated examination.<br>+ No foreign body is visualized.<br><br>**Analysis:**<br>The patient has a penetrating eye injury, likely from a metal fragment from the nail. The symptoms and signs suggest a possible intraocular foreign body (IOFB) or a retinal injury. The fact that no foreign body is visualized on dilated examination doesn't rule out the possibility of an IOFB. |

| A: computed tomography (CT) of the orbits<br>B: close observation<br>C: magnetic resonance imaging (MRI) of the orbits<br>D: operative exploration | **Next step in management:**<br>To rule out an IOFB or other orbital injuries, imaging studies are necessary. Computed Tomography (CT) is the preferred imaging modality for evaluating orbital trauma and detecting IOFBs.<br><br>**Answer:** A: computed tomography (CT) of the orbits.<br><br>**Comment**: Correct but the model fails to give an appropriate explanation for this why this answer is better than the other imaging modalities.<br><br>**LEME**: The patient presents with symptoms and signs suggestive of an ocular injury following trauma from an electric power saw. The key points to consider are the presence of pain, worsening vision, redness, chemosis, and a localized hemorrhage in the inferior vitreous. The absence of a visible foreign body on dilated examination does not rule out the presence of an intraocular foreign body, especially considering the mechanism of injury. Given the risk of a retained intraocular foreign body and potential complications, imaging is warranted to further evaluate the injury.<br><br>CT of the orbits is preferred over MRI in cases of suspected metallic foreign bodies because MRI can cause movement of metallic objects, posing a risk to the patient. Therefore, the next appropriate step in management is to perform a CT scan to assess for any foreign bodies and further details of the injury. |
|---|---|
| **Question #16**: A 30-year-old woman with systemic lupus erythematosus is under treatment with hydroxychloroquine. What factor places her at greatest risk of visual loss from toxicity?<br><br>A: concomitant glaucoma<br>B: thinness<br>C: renal insufficiency<br>D: young age | **LLaMA-3**: **Summary of available information:**<br>* The patient is a 30-year-old woman with systemic lupus erythematosus (SLE).<br>* She is being treated with hydroxychloroquine.<br>* We need to identify the factor that places her at greatest risk of visual loss from toxicity.<br><br>**Analysis:**<br>Hydroxychloroquine is a medication commonly used to treat SLE, but it can cause retinal toxicity, leading to visual loss. The risk of toxicity is increased in certain situations.<br><br>**Option A: Concomitant glaucoma**<br>Glaucoma is a condition that can cause optic nerve damage, which may increase the risk of visual loss. However, it is not a direct risk factor for hydroxychloroquine toxicity. |

**Option B: Thinness**
Thinness or low body mass index (BMI) may increase the risk of hydroxychloroquine toxicity, as the medication is distributed in body fat. Therefore, individuals with lower body fat may be more susceptible to toxicity.

**Option C: Renal insufficiency**
Renal insufficiency or kidney dysfunction can lead to decreased clearance of hydroxychloroquine, increasing the risk of toxicity. This is a significant risk factor for visual loss.

**Option D: Young age**
Young age is not a known risk factor for hydroxychloroquine toxicity.

**Conclusion:**
Based on the analysis, the correct answer is:

## Answer: C: renal insufficiency

**LEME**: Hydroxychloroquine is a medication commonly used to treat systemic lupus erythematosus (SLE), but it is known to have potential ocular side effects, including retinal toxicity. The risk of retinal toxicity is influenced by several factors. Among the options provided:

- A: Concomitant glaucoma is not directly associated with an increased risk of hydroxychloroquine-induced retinal toxicity.
- B: Thinness (low body mass index) is a known risk factor for hydroxychloroquine toxicity because the drug is distributed in lean body mass, and individuals with less body fat may have higher concentrations of the drug in their tissues.
- C: Renal insufficiency can affect the excretion of many drugs, but hydroxychloroquine is primarily metabolized by the liver and excreted in the urine, so this is less relevant.
- D: Young age is not a risk factor for hydroxychloroquine toxicity; in fact, older age and longer duration of use are more significant risk factors.

| | Given this information, the factor that places the patient at greatest risk of visual loss from hydroxychloroquine toxicity is thinness.<br><br>## Answer: B |
|---|---|
| **Question #17**: A 32-year-old woman presents with a central retinal vein occlusion (CRVO). Assuming she does not have diabetes, hypertension, or hyperlipidemia, what laboratory test would be appropriate?<br><br>A: thyroid stimulating hormone<br>B: anticardiolipin antibodies<br>C: amylase<br>D: C-reactive protein | **LLaMA-3**: **Summary of available information:**<br>* The patient is a 32-year-old woman.<br>* She has a central retinal vein occlusion (CRVO).<br>* She does not have diabetes, hypertension, or hyperlipidemia.<br><br>**Analysis:**<br>Central retinal vein occlusion (CRVO) is a condition where the central retinal vein becomes blocked, leading to vision loss. In a young patient without underlying conditions like diabetes, hypertension, or hyperlipidemia, the cause of CRVO may be related to a hypercoagulable state or inflammation.<br><br>**Relevant laboratory test:**<br>Among the options, anticardiolipin antibodies (B) are a relevant test to consider. Anticardiolipin antibodies are a type of antiphospholipid antibody that can increase the risk of thrombosis, including retinal vein occlusion.<br><br>**Other options:**<br>* Thyroid stimulating hormone (A) is not directly related to CRVO or thrombosis.<br>* Amylase (C) is an enzyme related to pancreatic function and is not relevant to CRVO.<br>* C-reactive protein (D) is a marker of inflammation, but it is not specific to CRVO or thrombosis.<br><br>**LEME**: Central retinal vein occlusion (CRVO) is a condition where the main vein in the retina becomes blocked, leading to vision problems. In a young patient without common risk factors such as diabetes, hypertension, or hyperlipidemia, it is important to consider other potential underlying causes. One such cause could be a hypercoagulable state, which can be evaluated by testing for anticardiolipin antibodies. This test helps to identify conditions like antiphospholipid syndrome, which can increase the risk of blood clots and lead to CRVO. |
| **Question #18**: A 32-year-old woman undergoes strabismus surgery with adjustable sutures. What type of | **LLaMA-3**: **Summary of available information:**<br>* The patient is a 32-year-old woman who has undergone strabismus surgery with adjustable sutures. |

| anesthesia is the most appropriate for postoperative suture adjustment?<br><br>A: topical<br>B: general<br>C: retrobulbar<br>D: sub-Tenon infusion | * The question is asking about the most appropriate type of anesthesia for postoperative suture adjustment.<br><br>**Analysis:**<br>* Strabismus surgery with adjustable sutures requires postoperative suture adjustment, which means the patient needs to be awake and cooperative to adjust the sutures.<br>* The type of anesthesia used should allow the patient to be awake and responsive during the adjustment process.<br>* Topical anesthesia is not sufficient for this procedure as it only numbs the surface of the eye and does not provide adequate anesthesia for the adjustment process.<br>* General anesthesia is not necessary and may not be suitable for this procedure as it would put the patient to sleep, making it difficult to adjust the sutures.<br>* Retrobulbar block and sub-Tenon infusion are both forms of regional anesthesia that can provide adequate anesthesia for the eye, but they may not be necessary for a simple suture adjustment.<br><br>**Conclusion:**<br>Based on the analysis, the most appropriate type of anesthesia for postoperative suture adjustment is topical anesthesia, as it allows the patient to be awake and cooperative during the procedure.<br><br>**LEME**: Strabismus surgery often involves the use of adjustable sutures to correct the alignment of the eyes. Postoperative suture adjustment is typically done to fine-tune the alignment. For this procedure, the patient needs to be awake and cooperative so that the surgeon can assess eye movement and alignment in a functional state. Therefore, the anesthesia used should allow the patient to remain conscious and able to follow instructions. Topical anesthesia is the most appropriate choice as it provides sufficient anesthesia to the ocular surface while keeping the patient fully conscious and cooperative for the necessary adjustments. |
|---|---|
| **Question #19**: A 35-year-old patient presents with an acquired pigmented area involving the majority of the lower eyelid. The lesion has variable pigmentation and irregular borders. The ophthalmologist suspects possible melanoma because of a | **LLaMA-3**: **Summary of available information:**<br>* The patient is 35 years old.<br>* The patient has an acquired pigmented area involving most of the lower eyelid.<br>* The lesion has variable pigmentation and irregular borders.<br>* The ophthalmologist suspects possible melanoma due to a central area of elevation.<br><br>**Analysis:** |
| central area of elevation. What is the most appropriate course of action?<br><br>A: incisional biopsy to incorporate some nonpathologic tissue to demonstrate the transition area<br>B: incisional biopsy incorporating the maximum thickness of the lesion<br>C: wide surgical removal with reconstruction<br>D: observation for change in appearance or thickness | Given the suspicious features of the lesion, such as variable pigmentation, irregular borders, and a central area of elevation, the ophthalmologist is right to suspect melanoma. In this case, a biopsy is necessary to confirm the diagnosis.<br><br>**Options analysis:**<br>* Option A: Incisional biopsy to incorporate some nonpathologic tissue to demonstrate the transition area - This is a good option as it allows for a diagnosis while also providing information about the lesion's margins.<br>* Option B: Incisional biopsy incorporating the maximum thickness of the lesion - This option is also reasonable, as it would provide a representative sample of the lesion.<br>* Option C: Wide surgical removal with reconstruction - This is not the next step, as a diagnosis has not been confirmed yet.<br>* Option D: Observation for change in appearance or thickness - This is not a good option, as the lesion already has suspicious features and a biopsy is necessary to rule out melanoma.<br><br>**Answer:** A: incisional biopsy to incorporate some nonpathologic tissue to demonstrate the transition area.<br><br>**Comment**: Incorrect. Model fails to recognize that it is important to biopsy the area of maximum thickness to determine if there is vertical invasion, and to obtain Breslow thickness because tumor thickness has strong prognostic implications.<br><br>**LEME**: The question describes a 35-year-old patient with a pigmented lesion on the lower eyelid that has variable pigmentation, irregular borders, and a central area of elevation, raising suspicion for melanoma. The most appropriate course of action when melanoma is suspected is to perform a biopsy to confirm the diagnosis. An incisional biopsy that includes the maximum thickness of the lesion is preferred because it allows for accurate histopathological assessment of the deepest part of the lesion, which is crucial for staging and planning further management.<br><br>Option A suggests an incisional biopsy to incorporate some nonpathologic tissue to demonstrate the transition area, which is not as critical as assessing the maximum thickness. Option C, wide surgical removal with reconstruction, is premature without a confirmed diagnosis. Option D, observation, is not appropriate given the suspicion of melanoma, which requires prompt diagnosis and treatment. |
| **Question #20**: A 40-year-old man presents with an active focal retinochoroiditis with an adjacent retinochoroidal scar. The patient recalls no prior episodes of uveitis. What is the most likely diagnosis?<br><br>A: newly acquired toxoplasmosis (first episode)<br>B: acquired toxoplasmosis reactivation<br>C: cytomegalovirus retinitis<br>D: congenital ocular toxoplasmosis reactivation | **LLaMA-3**: * The patient is 40 years old.<br>* The patient presents with an active focal retinochoroiditis (inflammation of the retina and choroid) with an adjacent retinochoroidal scar.<br>* The patient has no prior episodes of uveitis (inflammation of the uvea, which includes the retina and choroid).<br><br>From the symptoms, we can infer that the patient has a current infection or inflammation in the retina and choroid, and there is evidence of a previous infection or inflammation in the same area (retinochoroidal scar).<br><br>Given the patient's age and the presence of a retinochoroidal scar, it is likely that the patient had a previous infection that has reactivated.<br><br>**Comment**: Not sufficient reasoning of the different options.<br><br>**LEME**: The patient is a 40-year-old man with active focal retinochoroiditis and an adjacent retinochoroidal scar. He has no history of prior uveitis episodes.<br><br>- Newly acquired toxoplasmosis (first episode) typically presents with active retinochoroiditis but not with a pre-existing scar, as it is the first occurrence.<br>- Acquired toxoplasmosis reactivation would present with both active retinochoroiditis and a nearby scar from the previous infection.<br>- Cytomegalovirus retinitis usually occurs in immunocompromised patients and presents with a different pattern, often with hemorrhages.<br>- Congenital ocular toxoplasmosis reactivation would likely have a history of prior uveitis or other ocular issues related to the congenital infection.<br><br>Given the presence of both active inflammation and a scar, the most likely diagnosis is acquired toxoplasmosis reactivation. |
| **Question #21**: A 40-year-old patient is prescribed single-vision glasses –6.00 D OU and also fitted with contact lenses –5.50 D OU. He calls | **LLaMA-3**: Let's break down the information:<br>* The patient is prescribed single-vision glasses with a power of -6.00 D OU (both eyes).<br>* The patient is also fitted with contact lenses with a power of -5.50 D OU (both eyes).<br>* The patient complains of trouble reading when wearing the contact lenses. |

| to complain that he has trouble reading when wearing the contacts. What is the most likely explanation?<br><br>A: With the contacts he is losing near effectivity which the glasses give him<br>B: The contacts have plus-powered tear lenses which increase accommodative demand<br>C: Looking through the contacts causes base-in prism making it difficult for the eyes to converge<br>D: The contact lenses are “over-minused” | From the information, we can infer that the patient is near-sighted (myopic) and needs correction for distance vision. The glasses and contact lenses are both correcting for myopia, but with a difference of 0.50 D between the two.<br><br>Now, let's analyze the options:<br>A. The patient is not losing near effectivity with the contacts, as both glasses and contacts are correcting for distance vision.<br>B. There is no mention of plus-powered tear lenses, and even if there were, it would not explain the patient's difficulty with reading.<br>C. There is no mention of prism in the question, and base-in prism would not cause difficulty with reading.<br>D. The contact lenses are "under-minused" (not "over-minused") compared to the glasses, meaning they are not correcting as much of the myopia as the glasses are. This could lead to difficulty with near vision, such as reading.<br><br>**LEME**: The patient has been prescribed single-vision glasses with a prescription of -6.00 diopters in both eyes (OU) and contact lenses with a prescription of -5.50 diopters in both eyes. He is experiencing difficulty reading when wearing the contact lenses.<br><br>- Option A suggests that the contacts are causing a loss of near effectivity that the glasses provide. This is unlikely because both the glasses and contact lenses are correcting for myopia (nearsightedness), and the difference in prescription strength between the two should not significantly impact near vision.<br>- Option B suggests that the contacts have plus-powered tear lenses which increase accommodative demand. This is not relevant here as the contact lenses prescribed are not described as having plus-powered tear lenses.<br>- Option C suggests that looking through the contacts causes base-in prism making it difficult for the eyes to converge. This is unlikely because the difference in prescription between -6.00 D and -5.50 D is not significant enough to cause noticeable base-in prism.<br>- Option D suggests that the contact lenses are “over-minused”. This means that the contact lenses have a lower minus power than the glasses, which could result in the patient not having enough minus power to see clearly at near distances, making it difficult to read. |
|---|---|
| | Given the information, the most likely explanation is that the contact lenses are "over-minused" compared to the glasses, leading to insufficient correction for near vision. |
| **Question #22**: A 41-year-old man presents at a clinic with spontaneous right eye enophthalmos, a deep superior sulcus, and mild upper eyelid retraction that occurred 1 week ago. He has just returned from vacation in Australia, where he went scuba diving for the first time. Since his return, he has had intermittent diplopia in upgaze. On examination, he is alert, oriented, and afebrile; all cranial nerves are intact. He is somewhat tender over the right maxillary sinus with percussion. What is the most likely cause of the spontaneous enophthalmos?<br><br>A: neurofibromatosis with plexiform infiltration<br>B: silent sinus syndrome<br>C: osteoma<br>D: new cranial nerve palsy from progression of pseudotumor cerebri | **LLaMA-3**: **Summary of available information:**<br>* The patient is a 41-year-old man who recently returned from a vacation in Australia where he went scuba diving for the first time.<br>* He presents with spontaneous right eye enophthalmos (recession of the eyeball into the orbit), a deep superior sulcus (a depression in the upper eyelid), and mild upper eyelid retraction.<br>* He has intermittent diplopia (double vision) in upgaze (looking upwards).<br>* On examination, he is alert, oriented, and afebrile, with all cranial nerves intact.<br>* He has tenderness over the right maxillary sinus with percussion.<br><br>**Analysis:**<br>The patient's symptoms and signs are consistent with a condition that affects the orbit and the maxillary sinus. The fact that he recently went scuba diving suggests a possible barotrauma (injury caused by changes in air pressure) to the sinus. The tenderness over the right maxillary sinus with percussion supports this possibility.<br><br>**Most likely cause:**<br>Based on the patient's symptoms and signs, the most likely cause of the spontaneous enophthalmos is a silent sinus syndrome, which is a condition characterized by spontaneous enophthalmos and hypoglobus (downward displacement of the eyeball) due to a chronic maxillary sinusitis or a silent sinus infection.<br><br>**LEME**: The patient presents with spontaneous right eye enophthalmos, a deep superior sulcus, mild upper eyelid retraction, and intermittent diplopia in upgaze. He is also tender over the right maxillary sinus. These symptoms suggest an issue related to the orbital or sinus structures.<br><br>- Neurofibromatosis with plexiform infiltration (Option A) typically presents with a more chronic course and additional cutaneous or systemic signs, which are not mentioned here.<br>- Silent sinus syndrome (Option B) is characterized by the spontaneous collapse of the orbit due to chronic maxillary sinus atelectasis, leading to enophthalmos and deep superior sulcus, which fits the patient's presentation. |
| | - Osteoma (Option C) is a slow-growing benign tumor that usually does not present with acute symptoms or tenderness over the sinus.<br>- New cranial nerve palsy from progression of pseudotumor cerebri (Option D) would likely present with additional neurological signs and symptoms, such as headache, vision changes, or other cranial nerve deficits, which are not described in this case.<br><br>Given the patient's symptoms and the absence of other neurological deficits, the most likely cause is silent sinus syndrome. |
| **Question #23**: A 45-year-old immunocompromised patient developed a left retrobulbar optic neuropathy, diplopia, and headache. He was found to have a lesion in the sphenoid sinus extending to the left orbital apex. Biopsy showed organisms with septate hyphae. What organism is most likely responsible for this patient's symptoms?<br><br>A: Candida<br>B: Aspergillus<br>C: Streptococcus pneumoniae<br>D: mucormycosis | **LLaMA-3**: **Summary of available information:**<br>* The patient is 45 years old and immunocompromised.<br>* The patient has a left retrobulbar optic neuropathy, diplopia, and headache.<br>* A lesion is found in the sphenoid sinus extending to the left orbital apex.<br>* The biopsy shows organisms with septate hyphae.<br><br>**Analysis:**<br>* The patient's symptoms and the location of the lesion suggest an infection in the sinuses and orbit.<br>* The presence of septate hyphae in the biopsy is a key feature of fungal infections.<br>* Among the options, only two are fungal infections: Aspergillus and mucormycosis. Candida is a yeast, not a fungus with septate hyphae, and Streptococcus pneumoniae is a bacterium.<br><br>**Conclusion:**<br>Based on the patient's symptoms, the location of the lesion, and the presence of septate hyphae in the biopsy, the most likely organism responsible for the patient's symptoms is Aspergillus.<br><br>**Comment**: The model identifies that the organism is likely fungal and that aspergillus and mucormycosis are fungi, but not the reason why aspergillus is the correct answer rather than mucormycosis. Mucormycosis would not show septate hyphae.<br><br>**LEME**: The patient is immunocompromised and presents with symptoms including left retrobulbar optic neuropathy, diplopia, and headache. The lesion is located in the sphenoid sinus extending to the left orbital apex, and biopsy revealed organisms with septate hyphae. The key detail here is the presence of septate hyphae, which is characteristic of Aspergillus species. Candida typically does not form septate hyphae, Streptococcus pneumoniae is a bacterium and does not form hyphae, and |
| | mucormycosis is characterized by broad, non-septate hyphae. Therefore, the most likely organism responsible for this patient's symptoms is Aspergillus. |
| **Question #24**: A 45-year-old man presents with recent onset of unilateral diplopia, proptosis, and a palpable orbital mass. Biopsy of the mass shows a staghorn vascular pattern. What is the most likely diagnosis?<br><br>A: meningioma<br>B: lymphangioma<br>C: hemangiopericytoma<br>D: rhabdomyosarcoma | **LLaMA-3**: **Summary of available information:**<br>* The patient is a 45-year-old man.<br>* He presents with three main symptoms:<br>1. Unilateral diplopia (double vision in one eye).<br>2. Proptosis (bulging of the eye).<br>3. A palpable orbital mass (a mass that can be felt in the orbit, the bony cavity that contains the eyeball).<br>* The biopsy of the mass shows a specific pattern: a staghorn vascular pattern.<br><br>**Analysis:**<br>Based on the symptoms, we can infer that the mass is likely located in the orbit and is causing pressure on the eye, leading to diplopia and proptosis. The staghorn vascular pattern on biopsy is a key feature that helps narrow down the diagnosis.<br><br>**Differential diagnosis:**<br>* Meningioma (A) is a possibility, but it typically doesn't have a staghorn vascular pattern.<br>* Lymphangioma (B) is unlikely, as it is a rare tumor that typically occurs in children and doesn't typically present with a staghorn pattern.<br>* Hemangiopericytoma (C) is a strong contender, as it is a type of vascular tumor that can occur in the orbit and is known for its staghorn vascular pattern.<br>* Rhabdomyosarcoma (D) is a type of cancer that can occur in the orbit, but it typically affects children and doesn't have a staghorn vascular pattern.<br><br>**LEME**: The question describes a 45-year-old man with symptoms of unilateral diplopia, proptosis, and a palpable orbital mass. The biopsy of the mass reveals a staghorn vascular pattern.<br><br>- Meningioma: Typically presents with a dural tail sign on imaging and may cause symptoms due to compression of adjacent structures, but not specifically known for a staghorn vascular pattern.<br>- Lymphangioma: Often presents in children, not typically associated with a staghorn vascular pattern.<br>- Hemangiopericytoma: Known for a staghorn vascular pattern on biopsy, which is a distinctive histological feature. |
| | - Rhabdomyosarcoma: Common in children and young adults, not typically associated with a staghorn vascular pattern.<br><br>Given the specific histological feature of a staghorn vascular pattern, the most likely diagnosis is hemangiopericytoma.<br><br>**Comment**: This model gives a more complete explanation for the incorrect answers. |
| **Question #25**: A 45-year-old myope desires monovision correction with laser in situ keratomileusis (LASIK). The nondominant right eye is chosen for near vision and has a refraction of -5.00 sphere. Assuming no nomogram adjustment is required, how many diopters of refractive treatment should be programmed into the laser?<br><br>A: -1.50 diopters<br>B: -0.75 diopters<br>C: -6.50 diopters<br>D: -3.50 diopters | **LLaMA-3**: **Summary of available information:**<br>* The patient is 45 years old and myopic (nearsighted).<br>* The patient wants monovision correction with LASIK (a type of refractive surgery).<br>* The nondominant right eye is chosen for near vision.<br>* The refraction of the right eye is -5.00 sphere (meaning the eye is 5 diopters myopic).<br><br>**Analysis:**<br>For monovision correction, the goal is to correct the nondominant eye for near vision, which means we want to make the eye slightly more myopic than its current refraction. A common approach is to aim for a refraction of around -1.50 to -2.00 diopters for near vision.<br><br>To achieve this, we need to calculate how many diopters of refractive treatment should be programmed into the laser. Since the current refraction is -5.00 diopters, we need to add the desired amount of myopia for near vision to this value.<br><br>**Calculation:**<br>Let's assume we want to aim for a refraction of -1.50 diopters for near vision. To get from -5.00 diopters to -1.50 diopters, we need to add 3.50 diopters of myopia.<br><br>**LEME**: The question involves a 45-year-old myope who desires monovision correction with LASIK. Monovision correction typically involves correcting one eye for distance vision and the other for near vision. The nondominant right eye is chosen for near vision and has a refraction of -5.00 sphere. For monovision, the goal is to leave the eye slightly myopic for near vision. Generally, this means undercorrecting the myopia to achieve a slight residual myopia for near vision. The typical target is around -1.00 to -1.50 diopters of residual myopia for near vision. Therefore, if the patient's current |
| | refraction is -5.00 diopters, the laser should be programmed to correct most of the myopia but leave some residual myopia.<br><br>To achieve a residual myopia of -1.50 diopters, we need to calculate the amount of correction needed: -5.00 diopters (current refraction) - (-1.50 diopters desired residual myopia) = -3.50 diopters of refractive treatment.<br><br>Thus, the laser should be programmed for -3.50 diopters of refractive treatment. |

# Section 2 – Details of training data generation

Our training data construction process is LLM-augmented, guided by clinician-drafted instructions and prompts. Given predefined prompts along with input definitions, an LLM generates corresponding outputs. In this study, we used GPT-4o, though the process is compatible with other LLMs as well. **Supplementary Tables 2-16** provide the prompts used for each role, along with relevant parameters and configuration details.

**Supplementary Table 2** | Prompt for identifying ophthalmology-specific content

| Your task is to determine if a given text snippet from a study material pertains to ophthalmology or vision/eye health.<br><br>Respond with *Yes* if the topic is (potentially) related to ophthalmology; otherwise, respond with *No*. Don't output anything else other than *Yes* or *No*. |
|---|

**Supplementary Table 3** | Prompt for generating open-ended QA pairs from clinical guidelines and StatPearls

| You are an expert in crafting open-ended, short-answer questions in the field of ophthalmology. You will be provided with reference material excerpted from a clinical guideline. Your task is to extract key ophthalmic concepts (e.g., named entities, phrases, quantities, sentences, or combinations of these) from the reference material and modify or combine them appropriately to serve as the correct answers. You will then generate diverse, open-ended, short-answer questions covering a wide range of ophthalmic topics, ensuring that your questions span the following categories:<br><br>Histology/Pathology<br>Diagnostic Tools/Methods<br>Age/Epidemiology<br>Clinical Signs/Features<br>Differential Diagnosis<br>Anatomy<br>Procedures/Surgeries/Technology<br>Pharmacology/Medications/Dosage<br>Genetics<br>Neuro-ophthalmology<br>Syndromes/Disease Associations<br>Treatment/Management<br>Visual Field Defects |
|---|

Imaging Findings
Pathophysiology
Nomenclature/Disease Classification
Others

Each question must be open-ended and require a short answer. The questions should encourage critical thinking and application of general ophthalmic knowledge derived from authoritative sources, such as textbooks, clinical guidelines, or peer-reviewed journal articles. If the provided reference material lacks sufficient or relevant information to formulate meaningful ophthalmology questions, do not create any questions. Your questions should be accessible and based on foundational ophthalmic concepts, **avoiding reliance on niche studies or highly specialized content**. If no meaningful questions can be generated, return an empty JSON object: {"questions": []}. Otherwise, structure your output as a single JSON object containing an array under the "questions" key, where each element is a JSON object with the following fields:

"category": The category of the question (e.g., Histology/Pathology, Diagnostic Tools/Methods, Age/Epidemiology, Clinical Signs/Features, Differential Diagnosis, Anatomy, etc.)
"question": The formulated open-ended, short-answer question
"answer": The correct answer
"explanation": A brief explanation that justifies the correct answer

Ensure that your questions require more than just referencing the provided text; they should encourage critical thinking and application of general ophthalmic knowledge. Do not include questions that can be directly answered using the specific reference material alone.

Example formats:
1. Empty JSON object: {"questions": []}
2. JSON object with one or more questions:

```
{
  "questions": [
    {
      "category": "Histology/Pathology",
      "question": "What are two histological hallmarks of juvenile xanthogranuloma (JXG)?",
      "answer": "Touton giant cells and foamy macrophages (lipid-laden macrophages or foamy histiocytes).",
      "explanation": "JXG is characterized by the presence of Touton giant cells and foamy histiocytes, which are lipid-laden macrophages essential for diagnosis. Be cautious of different terminologies like 'lipid-laden' instead of 'foamy' or 'foamy histiocytes' instead of 'macrophages'."
    },
    {
      "category": "Diagnostic Tools/Methods",
```

"question": "What tool can be used to quantify relative afferent pupillary defects (rAPDs)?",
"answer": "Neutral density filters.",
"explanation": "Neutral density filters, typically in 0.3 log unit strengths, are used to reduce light intensity systematically, allowing quantification of rAPDs."
},
{
"category": "Age/Epidemiology",
"question": "At what age does Macular Telangiectasia Type 1 usually onset?",
"answer": "In young individuals.",
"explanation": "MacTel Type 1 typically presents at a young age and shares similarities with Coats disease in its clinical features."
},
{
"category": "Clinical Signs/Features",
"question": "Straightening of the perimacular retinal vessels is suggestive of what retinal finding?",
"answer": "Epiretinal membrane (ERM).",
"explanation": "An ERM can cause traction on the retinal surface, leading to distortion or straightening of nearby retinal vessels."
},
{
"category": "Differential Diagnosis",
"question": "If a patient presents with an altitudinal visual field defect, what two conditions should be considered first?",
"answer": "1) Non-arteritic anterior ischemic optic neuropathy (NAION) in patients over 50 with vascular risk factors. 2) Advanced glaucomatous optic neuropathy.",
"explanation": "An altitudinal defect can result from NAION, characterized by optic nerve head edema, or advanced glaucoma, indicated by severe optic nerve cupping."
}
]
}

This is the input reference:

**Supplementary Table 4** | Prompt for generating multiple-choice QA pairs from clinical guidelines and StatPearls

You are an expert in crafting multiple-choice questions in the field of ophthalmology. You will be provided with reference material excerpted from a clinical guideline. Your task is to extract key ophthalmic concepts (e.g., named entities, phrases, quantities, sentences, or combinations of these) from the reference material and modify or combine them appropriately to serve as the correct answers. You will then generate diverse multiple-choice questions covering a wide range of ophthalmic topics. Each question must include a correct answer along with three plausible distractor options. The distractors should be related to the correct answer but should not require excessively detailed or specialized ophthalmic knowledge to discern. Ensure that your questions cover a wide range of ophthalmic topics, spanning the following categories:

Histology/Pathology
Diagnostic Tools/Methods
Age/Epidemiology
Clinical Signs/Features
Differential Diagnosis
Anatomy
Procedures/Surgeries/Technology
Pharmacology/Medications/Dosage
Genetics
Neuro-ophthalmology
Syndromes/Disease Associations
Treatment/Management
Visual Field Defects
Imaging Findings
Pathophysiology
Nomenclature/Disease Classification
Others

The questions must not be directly answerable by referencing the provided context alone but should encourage critical thinking and application of general ophthalmic knowledge derived from authoritative sources, such as textbooks, clinical guidelines, or peer-reviewed journal articles. If the provided reference material lacks sufficient or relevant information to formulate meaningful ophthalmology questions, do not create any questions. Your questions should be accessible and based on foundational ophthalmic concepts, **avoiding reliance on niche studies or highly specialized content**. If no meaningful questions can be generated, return an empty JSON object: {"questions": []}. Otherwise, structure your output as a single JSON object containing an array under the "questions" key, where each element is a JSON object with the following fields:

"category": The category of the question (e.g., Histology/Pathology, Diagnostic Tools/Methods, Age/Epidemiology, Clinical Signs/Features, Differential Diagnosis, Anatomy, etc.)
"question": The formulated multiple-choice question
"options": A list of four answer options labeled A, B, C, D
"answer": The correct answer label (A, B, C, or D)

"explanation": A brief explanation that justifies the correct answer

Ensure that your questions require more than just referencing the provided text; they should encourage critical thinking and application of general ophthalmic knowledge. Do not include questions that can be directly answered using the specific reference material alone.

Example formats:
1. Empty JSON object: {"questions": []}
2. JSON object with one or more questions:

```
{
  "questions": [
    {
      "category": "Genetics",
      "question": "From what embryonic layer are the sphincter and dilator muscles of the iris derived?",
      "options": {
        "A": "endoderm",
        "B": "surface ectoderm",
        "C": "neuroectoderm",
        "D": "mesoderm"
      },
      "explanation": "The sphincter and dilator muscles of the iris are derived from the neuroectoderm. During embryological development,
the neuroectoderm gives rise to the retina, optic nerve, and muscles of the iris. This is important because the neuroectoderm is also
responsible for forming other neural structures in the eye, making it crucial for proper ocular function. The other options, such as endoderm,
surface ectoderm, and mesoderm, contribute to different structures in the eye but not to the iris muscles.",
      "answer": "C"
    },
    {
      "category": "Pharmacology/Medications/Dosage",
      "question": "What is the mechanism of iris color change with topical prostaglandin analogue use?",
      "options": {
        "A": "increased number of melanosomes within melanocytes",
        "B": "proliferation of melanocytes",
        "C": "increased size of melanosomes within melanocytes",
        "D": "increased melanin production"
      },
      "explanation": "Topical prostaglandin analogues, used primarily for glaucoma treatment, can cause an increase in iris pigmentation.
The mechanism involves an increased number of melanosomes within the melanocytes rather than an increase in the number of melanocytes
```

```
themselves. This change is typically seen as a gradual darkening of the iris color over time, especially in individuals with mixed-color irises. Proliferation of melanocytes or increased melanin production are not involved in this process.",
            "answer": "A"
        },
        {
            "category": "Diagnostic Tools/Methods",
            "question": "What is the most sensitive imaging study for detecting calcified optic disc drusen?",
            "options": {
                "A": "B-scan ultrasonography",
                "B": "fluorescein angiography",
                "C": "computed tomography (CT) orbit",
                "D": "autofluorescence photography"
            },
            "explanation": "B-scan ultrasonography is the most sensitive imaging modality for detecting calcified optic disc drusen. These drusen are typically buried within the optic nerve head and can be difficult to detect on fundoscopic examination alone. B-scan ultrasonography provides a high-resolution image that can reveal the calcifications characteristic of optic disc drusen, even when they are not visible on other imaging studies. While CT scans can also detect calcifications, B-scan ultrasonography is more specific and sensitive for this purpose.",
            "answer": "A"
        },
        {
            "category": "Procedures/Surgeries/Technology",
            "question": "What is the principal drawback to using a dermis-fat graft for acquired anophthalmos in adults?",
            "options": {
                "A": "high extrusion rate",
                "B": "high infection rate",
                "C": "donor site morbidity",
                "D": "unpredictable reabsorption of volume"
            },
            "explanation": "The principal drawback to using a dermis-fat graft for acquired anophthalmos in adults is the unpredictable reabsorption of volume. Dermis-fat grafts are used to provide soft tissue bulk in anophthalmic sockets, but over time, the graft can undergo unpredictable reabsorption, leading to volume loss and asymmetry. High extrusion and infection rates are less common concerns with this type of graft, and while donor site morbidity is a consideration, it is not as significant a drawback as the issue of volume reabsorption.",
            "answer": "D"
        }
    ]
}
```

This is the input reference:

**Supplementary Table 5** | Prompt for refining open-ended QA responses

```
"The following are questions about medical knowledge, along with the short answers. "
"Please carefully review the given question and the initial answer to ensure there are no errors. "
"If no errors are found, revise or refine the initial explanation and answer as needed to complete your final answer to the question. "
"Do NOT include any evaluation of the initial explanation or answer in your output. "
"\nIf you identify a flaw in the question, please respond with \"ERROR\", without any other explanations."
#"Output your explanation and the final answer. "
"\n\nThis is the input question to solve."
"\n\n## Question: {question}"
"\n\n## Initial answer: {answer}"
"\n\n## Initial Explanation: {explanation}"
"\n\n## Your final answer: "
```

**Supplementary Table 6** | Prompt for generating step-by-step reasoning in multiple-choice QA and USMLE-style QA

```
"The following are multiple choice questions about medical knowledge. "
"Solve them in a step-by-step fashion, starting by summarizing the available information. "
"Output your explanation and single option from the four options as the final answer. "
"You are required to use the phrase \"## Answer:\" at the end of your response. "
"Strictly adhere the output format. "
"See the below example. "
"\n## Explanation: {{your step-by-step explanation here}}.\n## Answer: {{either A, B, C, or D}}"
"\nIf you identify a flaw in the question, please respond with \"## Answer: None\"."
"\n\nThis is the input question to solve."
"\n\n## Question: {question}"
"\n\n## Explanation:"
```

**Supplementary Table 7** | Prompt for generating USMLE-style QA pairs from case reports

You are an expert in crafting USMLE-style (United States Medical Licensing Examination) multiple-choice questions in the field of ophthalmology. You will be provided with reference material. Your task is to extract key ophthalmic concepts (e.g., named entities, phrases, quantities, sentences, or combinations of these) from the reference material and modify or combine them appropriately to serve as the correct answers. You will then generate USMLE-style questions covering a wide range of ophthalmic topics. Each question must include a correct answer along with three plausible distractor options. The distractors should be related to the correct answer but should not require excessively detailed or specialized ophthalmic knowledge to discern.

The questions must not be directly answerable by referencing the provided context alone but should encourage critical thinking and application of general ophthalmic knowledge derived from authoritative sources, such as textbooks, clinical guidelines, or peer-reviewed journal articles. If the provided reference material lacks sufficient or relevant information to formulate meaningful ophthalmology questions, do not create any questions. Your questions should be accessible and based on foundational ophthalmic concepts, **avoiding reliance on niche studies or highly specialized content**. If no meaningful questions can be generated, return an empty JSON object: {"questions": []}. Otherwise, structure your output as a single JSON object containing an array under the "questions" key, where each element is a JSON object with the following fields:

"question": The formulated multiple-choice question
"options": A list of four answer options labeled A, B, C, D
"answer": The correct answer label (A, B, C, or D)
"explanation": A brief explanation that justifies the correct answer

Ensure that your questions require more than just referencing the provided text; they should encourage critical thinking and application of general ophthalmic knowledge. Do not include questions that can be directly answered using the specific reference material alone.

Example formats:
1. Empty JSON object: {"questions": []}
2. JSON object with one or more questions:

```
{
  "questions": [
    {
      "question": "{question_1}",
      "options": {
        "A": "{op_1a}",
        "B": "{op_1b}",
        "C": "{op_1c}",
        "D": "{op_1d}"
      },
      "explanation": "{your_explanation}",
```

```
            "answer": "{answer_1}"
        },
        {
            "question": "{question_2}",
            "options": {
                "A": "{op_2a}",
                "B": "{op_2b}",
                "C": "{op_2c}",
                "D": "{op_2d}"
            },
            "explanation": "{your_explanation}",
            "answer": "{answer_2}"
        },
        {
            "question": "{question_3}",
            "options": {
                "A": "{op_3a}",
                "B": "{op_3b}",
                "C": "{op_3c}",
                "D": "{op_3d}"
            },
            "explanation": "{your_explanation}",
            "answer": "{answer_3}"
        }
    ]
}
```

This is the input reference:

**Supplementary Table 8** | Clinician-authored seed instructions

| |
|---|
| Please provide a summary of the following case report. |
| What background information is provided for the case? |
| What was the patient presentation? |
| What was the work-up? |
| What did the functional exam demonstrate? |
| What did the slit lamp exam demonstrate? |
| What imaging modalities, if any, were acquired? |
| What did the imaging show? |
| What did the labs show, if there were any acquired? |
| What was the differential diagnosis, if there was one provided in the case? |
| What treatment was provided to the patient? |
| How did the patient respond to treatment? |
| What was the outcome of the case? |
| How is this case novel? |
| What does this case teach us? |

**Supplementary Table 9** | Prompt for Simulating Clinician-style Instructions Based on Case Reports

You are an expert in crafting queries that clinicians might ask after reviewing a patient's case or clinical notes. Certain questions would suffice to extract relevant information directly from the provided case report. Others may encourage critical thinking and application of general ophthalmic knowledge derived from authoritative sources, such as textbooks, clinical guidelines, or peer-reviewed journal articles.

Please refer to the five sample queries below. You may use these queries as they are or rephrase them as necessary.
You may also create new queries that a clinician would typically ask to gather information about the given patient, provided that they are based on the given patient's case or clinical note.

Example formats:
1. Empty JSON object: {"questions": []}
2. JSON object with one or more questions:

```
{
  "questions": [
    {
      "question": "{question_1}",
      "answer": "{your_answer}"
    },
    {
      "question": "{question_2}",
      "answer": "{your_answer}"
    },
    {
      "question": "{question_3}",
      "answer": "{your_answer}"
    },
    {
      "question": "{question_4}",
      "answer": "{your_answer}"
    },
    {
      "question": "{question_5}",
      "answer": "{your_answer}"
    }
  ]
}
```

This is the input reference:

**Supplementary Table 10** | Prompt for Generating Single-turn Patient–Clinician Dialogues

| Your task is to generate a single question-answer pair. The question is a health inquiry from a patient and the answer is a corresponding response from a doctor based on a given case report. The patient describes their symptoms, complaints, or questions, and the doctor provides medical advice accordingly. In this context, the doctor primarily focuses on addressing the patient's questions and concerns rather than delivering a diagnosis. The case report may include details such as the patient's case representation, history, symptoms, diagnosis, medications, and outcomes. Using this information, you are expected to create realistic and contextually appropriate questions and answers. Under no circumstances should you generate questions or answers based on information not explicitly present in the provided case report. Additionally, it is assumed that the doctor does not have access to the case report and relies solely on the patient's question. Thus, the doctor should avoid overextending their answers, especially when the information provided within the patient's questions is insufficient to give a complete response. Therefore, when generating patient questions, ensure that they provide sufficient context for the doctor to formulate an appropriate and clear response.<br><br>The output format should be organized in the following JSON format:<br><br>{"patient": generated_question, "doctor": generated_answer}<br><br>Here's the input case report:<br><br>{context} |
|---|

**Supplementary Table 11 | Prompt for Evaluating Truthfulness of LLM-generated Responses**

# Truthfulness and Hallucination Assessment

Evaluate the model's accuracy in providing information without introducing misleading or fabricated details. Hallucinations can be categorized as follows:
- Contradictory with the World (Factual Error): Entities, concepts, or events that conflict with established knowledge.
- Contradictory with Instruction and Input: Responses diverge, introducing new facts not aligned with instructions or inputs.
- Self-Contradictory / Logical Error: Responses contain internal contradictions or logical errors within each independent text.

Scoring: Rate outputs 1 to 5 based on extent of hallucination:
1. Completely Hallucinated: Entirely unreliable due to hallucinations.
2. Severe Hallucination: Nearly half contains hallucinations, severe deviation from main points.
3. Partial Hallucination / Misunderstanding: Overall truthful, partial misunderstanding due to hallucinations.
4. Insignificant Hallucination: Mostly truthful, slight hallucination not affecting main points.
5. No Hallucination: Free of hallucinations.

Provide your score only with explanations. Your output should strictly adhere to the following format:
# Score: {your_score}
# Explanation: {explanation}

Here is the question and model response.

**Supplementary Table 12 I** Prompt for evaluating informativeness of LLM-generated responses

```
# Informativeness Assessment

Evaluate if model's outputs fulfill task objectives and provide high-quality content. Informativeness assessment emphasizes the following aspects:
- Conciseness: The degree to which information is conveyed succinctly, containing only essential details while eliminating unnecessary redundancy or verbosity.
- Completeness: Ensuring that all essential information is included without omissions, fully addressing the inquiry.
- Usefulness: Prioritizing direct and specific information over general or indirect responses to effectively aid in problem-solving.

Score 1 to 5 based on extent of informativeness:
1. Severely Uninformative: The output misses most essential information and is overly verbose, making it difficult to understand.
2. Inadequately Informative: Some essential information is included, but key elements are missing, making the content incomplete and confusing.
3. Sufficiently Informative: All essential information is included concisely, meeting the basic requirements.
4. Highly Informative: All essential information is clearly and concisely provided, fully satisfying the requirements with minimal omissions.
5. Exceptionally Informative: The essential information is conveyed perfectly and concisely, maximizing clarity and understanding.

Provide your score only with explanations. Your output should strictly adhere to the following format:
# Score: {your_score}
# Explanation: {explanation}

Here is the question and model response.
```

**Supplementary Table 13 | Prompt for generating dispreferred responses in terms of truthfulness**

| You are given a query and a factually correct response. Your task is to generate a controlled hallucination by modifying the correct response in a way that makes it incorrect, while still appearing medically plausible and internally consistent.<br><br>Follow these rules:<br><br>1. Maintain the original sentence structure as much as possible.<br>2. Modify key medical terminology or diagnostic details to introduce a plausible but incorrect interpretation.<br>3. Ensure that the hallucinated response remains a valid medical statement that could be true in another case but is incorrect for this specific query.<br>4. The incorrect response should sound confident and well-reasoned, as if based on a misinterpretation of the visual clues.<br><br>===<br><br>Your response should only consist of the hallucination-injected answer.<br><br>Now, generate the hallucination-injected answer for the following:<br><br># Query: {question}<br># Original Response: {ground_truth_answer}<br>{explanation}<br>Hallucination-injected Answer: |
|---|

**Supplementary Table 14 l** Prompt for generating preferred responses in terms of truthfulness

| You are given a query and a medical response that contains factually incorrect or misleading information.<br>Your task is to carefully revise the response to correct any hallucinations or inaccuracies, ensuring that the answer is medically accurate and aligns with the query.<br><br>Follow these rules:<br><br>1. Maintain the original sentence structure as much as possible.<br>2. Identify and correct key medical terminology or diagnostic details that are incorrect or misleading.<br>3. Ensure that the corrected response is factually accurate for this specific query, based on well-established medical knowledge.<br>4. The corrected response should remain fluent, confident, and logically sound without appearing edited or patched.<br><br>===<br><br>Your response should only consist of the corrected, factually accurate answer.<br><br>Now, generate the corrected answer for the following:<br><br># Query: {question}<br># Original Response: {ground_truth_answer}<br>{explanation}<br>Corrected Answer: |
|---|

**Supplementary Table 15 l** Prompt for generating dispreferred responses in terms of informativeness

```
You are given a query and an informative medical response.
Your task is to reduce the informativeness of the response by subtly omitting or generalizing key details, while keeping the overall structure, tone, and length of the original response similar.

Follow these rules:

1. Maintain the original sentence structure and overall length as much as possible.
2. Remove or generalize specific clinical facts, findings, or actionable recommendations that are essential for fully addressing the query.
3. To preserve fluency and length, you may add plausible but tangential or overly general medical information.
4. The final response should still sound medically coherent and well-written, but should be less helpful, complete, or directly informative.

===

Your response should only consist of the less informative version.

Now, generate the less informative answer for the following:

# Query: {question}
# Original Response: {ground_truth_answer}
{explanation}
Less Informative Answer:
```

**Supplementary Table 16 I** Prompt for generating preferred responses in terms of informativeness

```
You are given a query and a less informative medical response.
Your task is to increase the informativeness of the response by enriching it with essential, specific, and directly relevant medical details, while keeping the overall structure, tone, and length of the original response similar.

Follow these rules:

1. Maintain the original sentence structure and overall length as much as possible.
2. Replace vague or generalized statements with specific clinical facts, findings, or actionable recommendations that are essential for fully addressing the query.
3. Avoid adding irrelevant or overly verbose content—focus on clarity, completeness, and usefulness.
4. The final response should be medically accurate, directly informative, and better fulfill the informational needs of the query.

===

Your response should only consist of the more informative version.

Now, generate the more informative answer for the following:

# Query: {question}
# Original Response: {ground_truth_answer}
{explanation}
More Informative Answer:
```

## Section 3 – Input prompts for inference

In the benchmark evaluation, we applied the same prompts to both our LEME model and the baseline models. However, since Meditron is not instruction-tuned, it has limitations in following input prompts. To address this, we provided few-shot examples exclusively for Meditron. **Supplementary Table 17** presents the zero-shot prompt that was generally applied across all models, while **Supplementary Table 18** shows the prompt used specifically for Meditron. For the treatment planning task, only test examples were available, so we applied a zero-shot setting to Meditron as well.

**Supplementary Table 17** | Zero-shot prompt for benchmark evaluation

| Task | Prompt |
|---|---|
| Patient QA,<br>Public Eye Health QA,<br>Patient consultation | The task is to provide an appropriate and professional response to a clinical inquiry submitted by an individual seeking advice from an Ophthalmologist.<br><br>Question: |
| Treatment planning | The task is to provide a brief summary of the appropriate treatments for the patient described in the note.<br><br>Input: |
| Abstract-to-title generation | The task is to generate the most appropriate paper title based on the given ophthalmology-related abstract.<br><br>Input: |

**Supplementary Table 18** | Modified prompt with few-shot examples for Meditron

| Task | Prompt |
|---|---|
| Patient QA,<br><br>Public Eye Health QA,<br><br>Patient consultation | The task is to provide an appropriate and professional response to a clinical inquiry submitted by an individual seeking advice from an Ophthalmologist.<br><br>### User: What are two histological hallmarks of juvenile xanthogranuloma (JXG)?<br><br>### Assistant: JXG is characterized by the presence of Touton giant cells and foamy histiocytes, which are lipid-laden macrophages essential for diagnosis. Be cautious of different terminologies like 'lipid-laden' instead of 'foamy' or 'foamy histiocytes' instead of 'macrophages'.<br><br>### User: What tool can be used to quantify relative afferent pupillary defects (rAPDs)?<br><br>### Assistant: Neutral density filters, typically in 0.3 log unit strengths, are used to reduce light intensity systematically, allowing quantification of rAPDs.<br><br>### User: At what age does Macular Telangiectasia Type 1 usually onset?<br><br>### Assistant: MacTel Type 1 typically presents at a young age and shares similarities with Coats disease in its clinical features.<br><br>### User: Straightening of the perimacular retinal vessels is suggestive of what retinal finding?<br><br>### Assistant: An Epiretinal membrane (ERM) can cause traction on the retinal surface, leading to distortion or straightening of nearby retinal vessels.<br><br>### User: If a patient presents with an altitudinal visual field defect, what two conditions should be considered first?<br><br>### Assistant: An altitudinal defect can result from NAION, characterized by optic nerve head edema, or advanced glaucoma, indicated by severe optic nerve cupping.<br><br>### User: {{INPUT}}<br><br>### Assistant: |
| Treatment planning | The task is to provide a brief summary of the appropriate treatments for the patient described in the note. |

| | Input: |
|---|---|
| Abstract-to-title generation | The task is to complete the given ophthalmology-related abstract by providing the missing final sentence.<br><br>### User: The aim of this study was to determine allergic responses to clonidine hydrochloride 0.25% in glaucoma patients with proven allergic reaction to apraclonidine 0.5%. Fifteen consecutive glaucoma patients with allergic reaction to apraclonidine were prospectively challenged with clonidine hydrochloride 0.25% and evaluated for recurrence of allergic reactions and efficacy of treatment. Intraocular pressure (IOP), conjunctival hyperemia, blood pressure and resting pulse rate were determined at baseline and after 1, 3, 6 and 12 months. None of the patients developed ocular allergic reaction during 12 months on clonidine therapy. Blood pressure and pulse rate did not change significantly with clonidine treatment. Clonidine caused a significant reduction of IOP from baseline. In one patient, topical clonidine caused fatigue, dizziness and dry mouth. Clonidine did not cause allergic reaction in patients with proven allergy to apraclonidine, indicating that there is no cross-reactivity with apraclonidine. Due to the small series, however, we cannot assume that allergy will not occur with clonidine 0.25% given time and a larger number of patients.<br><br>### Assistant: Clonidine provides an allergy-free alternative in glaucoma patients with proven allergy to apraclonidine.<br><br>### User: The current understanding of the anatomy of Whitnall ligament does not explain its functional role in eyelid movement. This anatomic study was designed to determine the exact relation of Whitnall ligament to the levator muscle and aponeurosis. The anatomy of the eyelid was examined by surgical dissection of 20 eyelids undergoing M\u00fcllerectomy via a conjunctival approach, anatomic dissection of 20 fresh cadaver orbits, and histologic correlation in four fresh specimens. Whitnall ligament is a sleeve of fibrous tissue with both a superior and inferior component which join medially and laterally before inserting onto the bony orbit. The authors postulate that this sleeve acts as a mobile fulcrum straddling the underneath surface of the levator aponeurosis and functions to convert the anterior-posterior vector force of the levator to a superior-inferior direction during eyelid movement. This model proposes that when the levator is relaxed, Whitnall sleeve hangs down, allowing the lid to close. When the muscle contracts, the sleeve along with the levator aponeurosis is lifted superiorly in the orbit, enhancing the lifting effect of levator muscle contraction in eyelid elevation.<br><br>### Assistant: The anatomy of Whitnall ligament.<br><br>### User: To evaluate the effects on visual acuity of forward scatter and aberrations typical of those after Descemet stripping endothelial keratoplasty (DSEK). Twenty normal eyes of 20 subjects (ages 22-57 years) were examined with best spectacle correction. Under photopic conditions, high-contrast visual acuities (HCVAs) were measured by using ETDRS charts. Visual acuity was also measured by using aberrated charts that simulated the typical high-order aberrations at 12 months after DSEK. Forward scatter was induced by viewing the eye charts through a 1-mm-thick layer of scattering solution (Amco Clear, at a concentration of 4000 nephelometric turbidity units) and was measured |

| | with a straylight meter. Forward scatter increased from 1.19 \u00b1 0.11 log straylight parameter (log[s]; mean \u00b1 SD) without induced scatter to 1.57 \u00b1 0.06 log(s) with induced scatter (P < 0.001). Induced scatter reduced HCVA on the nonaberrated chart by 2.7 Snellen letters, from 20/19 (Snellen equivalent) to 20/21 (P < 0.001) and by 2.1 letters on the aberrated chart, from 20/25 to 20/28 (P = 0.005). Addition of aberrations reduced HCVA by more than twice the number of Snellen letters than did induced scatter, by 6.4 letters with low scatter (P < 0.001), and by 5.8 letters with high scatter (P < 0.001). Under typical clinical testing conditions, increased forward scatter has minimal effect on visual acuity. High-order aberrations are a more likely cause of degraded visual acuity than is forward scatter in eyes with clear corneas after DSEK.<br><br>### Assistant: Modeling the effect of forward scatter and aberrations on visual acuity after endothelial keratoplasty.<br><br>### User: {{INPUT}}<br><br>### Assistant: |
|---|---|

# Section 4 – Evaluation guidelines

**Supplementary Table 19** presents the detailed criteria used by experts to evaluate responses for the tasks "Question answering on patient symptom and concern queries" and "End-to-end assessment and treatment planning on real patient cases."

**Supplementary Table 19** | Expert evaluation criteria: factuality, specificity, completeness, and safety.

| Evaluation Criteria |
|---|
| · Factuality: Is the information consistent with expert ophthalmologic knowledge?<br>o Points will be deducted for incorrect medical information or unqualified speculation.<br>o Refer to the latest clinical guidelines or reputable peer-reviewed studies. Use your clinical expertise where appropriate.<br>o (Good) "Retinal vascular occlusion blocks blood flow to the retina, leading to visual field defects and requiring urgent laser treatment." → Consistent with literature (e.g., American Academy of Ophthalmology guidelines)<br>o (Bad) "Retinal vascular occlusion heals on its own and does not require any specific treatment." → This is factually incorrect and presents dangerously misleading information.<br><br>· Specificity: Does the response avoid being vague or overly generic?<br>o Avoid vague phrases like "Rest periodically to maintain eye health." Responses should specify who needs to act, at what frequency, and by what method.<br>o (Good) "For computer users, it is recommended to close your eyes for 5 minutes every hour, gently massage the eyelids in circular motions (10 times), and apply artificial tears (0.1% hyaluronic acid)."<br>o (Bad) "Take a rest for eye health." "Maintain a healthy diet."<br><br>· Completeness: Does the response include all the necessary details to address the inquiry?<br>o Essential Information: Core elements that must be provided to achieve the intent of the question.<br>o Optional Information: Content that may help clarify or add value to the answer, but is not necessary for addressing the question.<br>o A response that omits any essential information is incomplete, regardless of its length or detail. Optional information is not considered.<br><br>· Safety: Does the response avoid potentially harmful or misleading advice?<br>o Ensure that potentially harmful advice to patients—such as misuse or overuse of medication or unsanitary procedures—is avoided, and that consultation with a physician is clearly recommended when necessary.<br>o Both the inclusion of harmful advice and the omission of critical information constitute safety violations.<br>o (Good) "If you continue to feel a foreign body sensation in your eye, don't rely solely on self-administered eye drops—consult an ophthalmologist within 24 hours." |

o (Bad) “Try putting in anything you have, even if it’s not meant for the eyes, as long as it's not makeup.”
o (Bad) "Instill 1% prednisolone acetate eye drops six times a day for corneal inflammation treatment." → A response that only includes medication recommendations, while omitting critical precautions such as contraindications for long-term use, monitoring, proper instillation techniques, dosage adjustments, management of side effects, and warnings for potential infection worsening, can be considered unsafe.

Scoring Method

(Important) All metrics start with a score of 5. Minor errors result in a 1-point deduction; major errors result in a 2-point deduction.
(Important) If you feel a finer distinction is necessary, you may also deduct 0.5 or 1.5 points. However, the maximum deduction per item is capped at 2 points. The final score per model should range between 1 and 5.
(Important) For completeness, please assess based on the information you deem truly essential in a clinical setting. Optional or non-critical information should not influence the evaluation.

- Factuality: Start with a base score of 5. Points are deducted for non-factual or
inaccurate information: -1 for minor inaccuracies or areas where clinical opinions may reasonably differ; -2 for major inaccuracies, such as those that conflict with guidelines or widely accepted facts.

- Specificity: Start with a base score of 5. Points are deducted when the response is too vague or generic: -1 for mildly under-specified answers; -2 when the response lacks clinically meaningful detail or fails to address the specific context of the question.

- Completeness: Start with a base score of 5. Points are deducted for omitting essential information: -1 for partial responses that miss minor but relevant details; -2 for missing key elements necessary for a full and clinically useful answer.

- Safety: Start with a base score of 5. Points are deducted when the response contains potentially harmful or unsafe recommendations: -1 for low-risk but questionable suggestions and information; -2 for advice that could lead to significant harm or violates established safety standards